\documentclass[twoside]{article}

\usepackage{graphicx}
\usepackage{algorithm}
\usepackage{amsmath}
\allowdisplaybreaks
\usepackage{amssymb}
\usepackage{appendix}
\usepackage{authblk}
\usepackage{bm}
\usepackage{booktabs}
\usepackage{enumitem}
\usepackage{hyperref}
\usepackage[utf8]{inputenc}
\usepackage{listings}
\usepackage{mathtools}
\usepackage{xcolor}
\usepackage{natbib}
\newtheorem{assumption}{Assumption}

\usepackage{enumitem,amssymb}
\usepackage[accepted]{aistats2025}
\begin{document}



\twocolumn[

\aistatstitle{Do Regularization Methods for Shortcut Mitigation Work As Intended?}

\aistatsauthor{ Haoyang Hong\And Ioanna Papanikolaou \And Sonali Parbhoo }
\vspace{0.25em}
\aistatsaddress{Imperial College London \And Imperial College London \And Imperial College London} ]

\begin{abstract}

Mitigating shortcuts, where models exploit spurious correlations in training data, remains a significant challenge for improving generalization. Regularization methods have been proposed to address this issue by enhancing model generalizability. However, we demonstrate that these methods can sometimes overregularize, inadvertently suppressing causal features along with spurious ones. In this work, we analyze the theoretical mechanisms by which regularization mitigates shortcuts and explore the limits of its effectiveness. Additionally, we identify the conditions under which regularization can successfully eliminate shortcuts without compromising causal features. Through experiments on synthetic and real-world datasets, our comprehensive analysis provides valuable insights into the strengths and limitations of regularization techniques for addressing shortcuts, offering guidance for developing more robust models.

\end{abstract}

%%%%%%%%%%%%%%%%%%%%%%%%%%%%%%%%%%%%%%%%%%%%%%%%%%%%%%%%%%%%%%%%%%%%%%%%%%%%
%%%%%%%%%%%%%%%%%%%%%%%%%%%%%%%%%%%%%%%%%%%%%%%%%%%%%%%%%%%%%%%%%%%%%%%%%%%%
%%%%%%%%%%%%%%%%%%%%%%%%%%%%%%%%%%%%%%%%%%%%%%%%%%%%%%%%%%%%%%%%%%%%%%%%%%%%
\section{Introduction}
% Machine learning (ML) models have demonstrated tremendous potential across several different applications and are increasingly being deployed in sensitive and high-stake environments to make important and life-changing decisions, such as hiring \citep{sam2023hiring}, criminal justice \citep{yacoby2022if}, healthcare \citep{balagopalan2024machine} and banking \citep{leo2019machine}. 
\vspace{-0.5em}
Despite successes of modern machine learning (ML), ML models frequently fail to generalize in scenarios where distribution shift occurs \citep{beery2018recognition,ilyas2019adversarial,subbaswamy2019preventing,arjovsky2019invariant} due to shortcut learning \citep{jabbour2020deep,geirhos2018generalisation,azulay2019deep}. Shortcut learning is defined as follows: \emph{``A predictor relies on input features that are easy to represent (i.e., shortcuts) and are predictive of the label
in the training data, but whose association with the label changes under relevant distribution shifts."}\citep{makar2022causally}. If left unresolved, models tend to exploit these spurious correlations between input attributes and output to produce high accuracy predictions on training data ~\citep{hermannfoundations}. However, when these models encounter out-of-distribution data (OOD) or data from a different distribution at test time, exploiting the incorrect information can negatively impact a model's generalization and robustness.

Several methods have been proposed to address a model's reliance on spurious attributes~\citep{ye2024spurious}. Some of these methods assume that the spurious attributes are known apriori and develop regularization strategies \citep{veitch2021counterfactual,makar2022causally,kaur2022modeling, wang2022learning}, optimization techniques \citep{sagawa2019distributionally} or data augmentation strategies \citep{kaushik2021explaining,joshi2021investigation} to explicitly make a model invariant to those specific attributes. \cite{kumar2024causal} do not explicitly assume an attribute is either spurious or not, and develop a method for regularizing the effect of attributes proportional to their \emph{average causal effect}. While \citet{kumar2024causal} is significantly more general than former approaches, it relies on several specific assumptions about the data generating process that may not hold in practice or may not be available in advance. 

Unlike previous works, this paper provides a comprehensive analysis (both theoretically and empirically) of the performance of regularization methods for shortcut mitigation and demonstrates that these methods can sometimes behave counterintuitively than indicated in existing literature. We identify several failure modes where issues remain unaddressed, particularly when shortcuts are unknown a priori or correlate with the outcome of interest. Specifically, we (1) derive the theoretical conditions for the success of these strategies and explore their practical limits; (2) demonstrate through synthetic and real experiments that spurious information can persist in concepts, allowing the model to exploit it for improved predictions even after regularization, especially when shortcuts correlate with input features or outputs; and (3) perform an in-depth analysis on strengths and limitations of regularization techniques for addressing shortcut learning and provide practical guidance and considerations for developing models that are robust to shortcuts.

% Unlike earlier works, in this paper we present a comprehensive analysis (both theoretically and empirically) of the performance of regularization methods for shortcut mitigation and demonstrate that these methods can sometimes behave counterintuitively than indicated in existing literature. In particular, we identify several failure modes of these approaches where the issues associated with the problem are not entirely addressed, particularly if shortcuts are unknown apriori or if these correlate with the outcome of interest. (1) We derive the precise theoretical conditions under which these strategies are likely to succeed and explore the limits of these in practice. (2) On a series of synthetic and real experiments, we demonstrate that spurious information can continue to be encoded in the concepts which can be exploited by the model to improve predictions even after the model has been regularized. Unfortunately this is common, particularly in cases where the shortcuts are correlated with input features other than the concepts. Finally,  we (3) perform an in-depth analysis on strengths and limitations of regularization techniques for addressing shortcut learning and provide practical guidance and considerations for developing models that are robust to shortcuts.
  
%%%%%%%%%%%%%%%%%%%%%%%%%%%%%%%%%%%%%%%%%%%%%%%%%%%%%%%%%%%%%%%%%%%%%%%%%%%%
%%%%%%%%%%%%%%%%%%%%%%%%%%%%%%%%%%%%%%%%%%%%%%%%%%%%%%%%%%%%%%%%%%%%%%%%%%%%
%%%%%%%%%%%%%%%%%%%%%%%%%%%%%%%%%%%%%%%%%%%%%%%%%%%%%%%%%%%%%%%%%%%%%%%%%%%%
\vspace{-0.3em}\section{Related work}
\vspace{-0.5em}\textbf{Model-based Methods for Shortcut Mitigation.}
Shortcuts occur as a result of distribution shift \citep{subbaswamy2019preventing,arjovsky2019invariant}. Invariant risk minimization \citep{arjovsky2019invariant} aims to make models invariant to changes in the data distribution across environments, encouraging reliance on stable, causal features as oppose to shortcuts. Group-based methods \citep{sagawa2019distributionally} model multiple groups of the training data based on the spurious attribute and optimize for worst group accuracy. This requires knowledge of the spurious variables apriori. Other methods such as \citep{liu2021just, creager2021environment} relax this assumption to require the spurious attribute only for validation data.  In contrast to these, our work considers how regularization methods perform for shortcut removal.

\textbf{Regularization Methods for Credible Models.}
Many works have considered augmenting the data to decorrelate shortcuts with data \citep{cubuk2018autoaugment} or regularizing the model to not consider the shortcut \citep{wang2022learning,veitch2021counterfactual, kaur2022modeling,zwaan2012relating}. However, these methods fail when shortcuts and features are correlated since without prior knowledge, these are indistinguishable. Hybrid methods such as \citet{zhang2018mitigating, ganin2016domain} combine model-based methods with regularization to consider reweighting the data or using domain-adversarial training to learn invariant representations and avoid shortcuts. These methods all assume shortcuts are known \emph{apriori}. \citet{zheng2022causally} consider automatically discovering spurious attributes by building risk invariant predictors that are additionally regularized to reduce variance. Further, \cite{pezeshki2021gradient} proposed using the network logits as regularization to achieve better generalization performance in binary classification problems. Unlike these works, we demonstrate that regularization methods can behave in counterintuitive ways than expected which could have significant impact on the overall influence of shortcuts.

\textbf{Causal Methods for Shortcut Mitigation.}
% Current approaches to mitigating shortcuts can typically be considered model-centric (e.g. invariant risk minimization, causal inference) or data-centric or a hybrid in nature. Invariant risk minimization \citet{arjovsky2019invariant} aim to make models invariant to changes in the data distribution across environments, encouraging reliance on stable, causal features as oppose to shortcuts. 
Causal methods such as \citet{scholkopf2021toward} explicitly leverage causal structures via causal representation learning methods to avoid relying on shortcuts. \citet{kumar2024causal} propose a technique for regularizing attributes proportional to their average causal effect. The authors use the fact that changing spurious attributes will not lead to a change in the task label and should thus have zero causal effect on the task label. However, the authors assume complete knowledge of the underlying causal structure of the data which may not always be available, particularly if certain key variables are unknown. 
%%%%%%%%%%%%%%%%%%%%%%%%%%%%%%%%%%%%%%%%%%%%%%%%%%%%%%%%%%%%%%%%%%%%%%%%%%%%
%%%%%%%%%%%%%%%%%%%%%%%%%%%%%%%%%%%%%%%%%%%%%%%%%%%%%%%%%%%%%%%%%%%%%%%%%%%%
%%%%%%%%%%%%%%%%%%%%%%%%%%%%%%%%%%%%%%%%%%%%%%%%%%%%%%%%%%%%%%%%%%%%%%%%%%%%
\vspace{-0.3em}\section{Setup and Preliminaries}
\vspace{-0.5em}

% The Concept Bottleneck Model (CBM) \citep{koh2020concept} is designed to train a model by mapping inputs to interpretable features that are known to influence the outputs. This design inherently mitigates shortcut effects. However, CBMs often underperform compared to end-to-end models \citep{havasi2022addressing}, largely due to the incomplete set of known concepts resulting from limited domain knowledge or insufficient data. To address these challenges, we augment CBM with an additional feature extractor to capture both causal and non-causal features. This enhancement not only compensates for the predictive performance gap but also ensures that regularization methods can be seamlessly integrated into the framework.
Concept Bottleneck Models (CBMs) \citep{koh2020concept} are designed to map inputs to interpretable features that are known to influence the outputs. This design can mitigate shortcut effects with the captured concepts but is not guaranteed to. Specifically, CBMs can often underperform compared to end-to-end models when there is an incomplete set of known concepts resulting from limited domain knowledge or insufficient data, or if the concepts themselves contain shortcut information \citep{havasi2022addressing}. Combining a CBM with a feature extractor to capture both causal and non-causal features may help compensate for the predictive performance gap and enables regularization methods to be seamlessly integrated into the framework, as has been demonstrated in \citet{wang2022learning}.

Consider a dataset $D = \{\bm{x}_i, \bm{c}_i, y_i\}_{i=1}^n$ consisting of $n$ samples, where $y_i \in \mathbb{R}$ represents the target output feature, and $\bm{x}_i \in \mathbb{R}^d$ are the input features. The vector $\bm{c}_i \in \mathbb{R}^c$ contains $c$ binary or continuous \textbf{known concepts} (denoted by $\bm{C}$), which are known to cause $y_i$ based on domain knowledge. A concept bottleneck model (CBM) \citep{koh2020concept} $f_{\theta_c}\circ f_{\theta_x}:X\rightarrow C\rightarrow Y$ parameterized by $\theta_c$ and $\theta_x$ can be trained accordingly by equation \ref{equ:cbm_loss}.
\begin{multline}\label{equ:cbm_loss}
      \hat{\theta_c}, \hat{\theta_x} = \arg\min_{\theta_c, \theta_x} \frac{1}{n}\sum_{i}^n[\mathcal{L}(f_{\theta_x}(\bm{x}_i), c_i) \\
    +\lambda_{cbm}\mathcal{L}(f_{\theta_c}\circ f_{\theta_x}(\bm{x}_i), y_i)]  
\end{multline}
where $\mathcal{L}$ is the loss function and $\lambda_{cbm}$ is the hyper-parameter that balances two losses. Beyond the pre-defined $\bm{C}$, we define the remaining concepts as \textbf{unknown concepts} ($\bm{U}$), assuming that $\bm{C}$ and $\bm{U}$ adhere to Assumption 1.

% Follow the setting in \citet{havasi2022addressing} and \citet{wang2022learning}, we abstract the unseen concepts as \textbf{unknown concepts} $U$. It is assumed that $C$ and $U$ follow:

\begin{figure}
    \centering
    \includegraphics[width=0.8\linewidth]{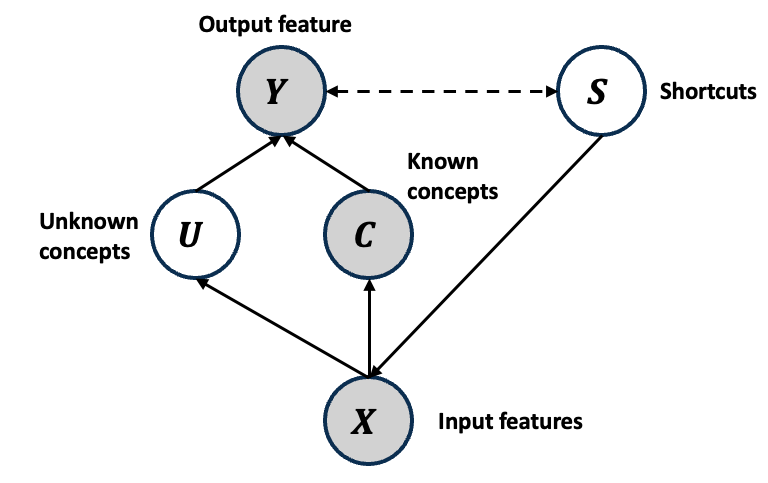}
    \caption{Causal graph for shortcut learning. $Y\in\mathbb{R}$ is the output feature. $\bm{X}\in\mathbb{R}^d$ represents the input features. $\bm{C}\in\mathbb{R}^c$ denotes the known concepts. Unknown concepts ($\bm{U}\in\mathbb{R}^u$) also cause $Y$ but are not observed or known. Shortcuts ($\bm{S}\in\mathbb{R}^s$) that have spurious correlation with $Y$ can be extracted from $\bm{X}$. $\bm{U}$ and $\bm{S}$ are mixed together. Observed variables are in gray. Dashed/solid edges represent correlation that is broken/unaffected under distribution shifts.}
    \label{fig:causal_graph}
\end{figure}

\begin{assumption}[Completeness of concepts]
\label{assumption1}
\textit{The known concepts $\bm{C}$ and unknown concepts $\bm{U}$ together constitute the complete set of features that causally influence the output feature $Y$.} 
\end{assumption} 
In practice, $\bm{U}$ can be derived from models pre-trained on unbiased datasets for similar tasks or large datasets like ImageNet~\citep{deng2009imagenet}, which capture general features that serve as unknown concepts. However, during feature extraction, there is a risk of identifying shortcut features ($\bm{S}$) that are spuriously correlated with the output feature $Y$ but not causally related. For instance, when inputting an image ($\bm{X}$) of a bird to determine whether it is a seabird or a landbird, known concepts ($\bm{C}$) such as the presence of wings or a beak can be extracted from the dataset. Unknown concepts ($\bm{U}$) are causally related features that are not measured in the dataset and can be extracted from pre-trained models. However, some extracted features, such as those representing the background color of the image, may be spurious concepts ($\bm{S}$). The causal relationships among all variables are illustrated in Figure~\ref{fig:causal_graph}.

After extracting $\bm{U}$ alongside the $\bm{S}$, we assume the linear relationships as outlined in Assumption 2~\eqref{assumption2}. This assumption is justified given that both $\bm{U}$ and $\bm{S}$ are in a latent space, meaning linear relationships are feasible, although this could also be relaxed (see supplement for details \ref{nonlinear_case}). 

\begin{assumption}[Linear relationship]
\label{assumption2}
    \textit{Known concepts ($\bm{C}\in\mathbb{R}^c$) and unknown concepts ($\bm{U}\in\mathbb{R}^u$) have a linear relationship with output feature $Y$, i.e.
\begin{align}Y&=\sum_{i=1}^c\beta_{ci}C_i + \sum_{j=1}^u\beta_{uj}U_j + \epsilon, \hspace{0.5em}\epsilon\sim\mathcal{N}(0,\sigma_{\epsilon}^2)
    % &=\bm{\beta_c}^T\mathbf{C} + \bm{\beta_u}^T\mathbf{U} + \epsilon, \hspace{0.5em}\epsilon\sim\mathcal{N}(0,\sigma_{\epsilon}^2)
\end{align}}
\end{assumption}

% Additionally, $S\in\mathbb{R}^s$ is correlated with $Y$ in the training dataset due to its correlation with $C$ and $U$.
% \begin{multline}
% S_k=\sum_{i=1}^c\delta_{cik}C_i+\sum_{j=1}^u\delta_{ujk}U_j+\epsilon_{sk},\hspace{0.5em}\epsilon_{sk}\sim\mathcal{N}(0,\sigma_{\epsilon_{sk}}^2)\\
% \forall k\in\{0,1,\cdots,s-1\}
% \end{multline}

Then the objective of mitigating shortcuts is equivalent to recovering $\beta_{ci},i\in\{1,2,\cdots,c\}$ and $\beta_{uj},j\in\{1,2,\cdots,u\}$ given input $\{\bm{C}, \bm{U}, \bm{S}\}$, while assigning $0$ to all shortcut features. Since $\bm{S}$ and $\bm{U}$ are mixed from unknown concepts extraction, distinguishing between shortcut and causally relevant features prior to model input is not feasible. Therefore, it is impossible to get rid of shortcuts before feeding the data to the model. Consequently, regularization techniques applied to this linear layer aim to effectively mitigate the influence of shortcuts. Formally, the  problem is defined as follows.

\textbf{Problem Setting.}\label{problem_setting} \textit{Given $\bm{X}\in\mathbb{R}^d$, $\bm{C}\in\mathbb{R}^c$ and $Y\in\mathbb{R}$ representing input features, known concepts, and output features respectively, train a CBM ($f_{\theta_c}\circ f_{\theta_x}$) based on $\bm{X}$, $\bm{C}$ and $Y$ following Equation~\eqref{equ:cbm_loss}. Apply a pre-trained model ($f_{\theta_{us}}: \bm{X}\rightarrow\{\bm{U}\in\mathbb{R}^u, \bm{S}\in\mathbb{R}^s\}$) to $\bm{X}$ for $\{\bm{U}, \bm{S}\}$ extraction. Assume $\bm{C}$ and the extracted $\bm{U}$, $\bm{S}$ satisfy Assumption 1 and 2. Then, based on $\bm{H}=\{\bm{C}, \bm{U}, \bm{S}\}\in\mathbb{R}^{h}$ where $h=c+u+s$, a linear model~\eqref{equ:linear_y} is trained by minimizing a loss function with regularization terms~\eqref{equ:regularization} to recover the true coefficients assigned to each variable. The targeted model is:
\begin{align}\label{equ:linear_y}
    \hat{Y}&=\sum_{i=1}^h\hat{\beta}_iH_i=\bm{\hat{\beta}}^T\bm{H}\\
    &=\bm{\hat{\beta_c}}^T\mathbf{C}+\bm{\hat{\beta_{us}}}^T[\mathbf{U}\hspace{0.5em}\mathbf{S}]\\
&=\bm{\hat{\beta_c}}^T\mathbf{C}+\bm{\hat{\beta_u}}^T\mathbf{U}+\bm{\hat{\beta_s}}^T\mathbf{S}\\
&=\bm{\hat{\beta_c}}^T(f_{\theta_x}(\bm{X}))+\left[\bm{\hat{\beta_u}}^T\hspace{0.5em}\bm{\hat{\beta_s}}^T\right]f_{\theta_{us}}(\bm{X})
\end{align}
The loss function with regularization is:
\begin{equation}\label{equ:regularization}
    loss=\mathcal{L}(Y, \hat{Y})+ \lambda_{reg}\mathcal{R}(\hat{\bm{\beta_c}}, \hat{\bm{\beta_u}}, \hat{\bm{\beta_s}})
\end{equation}
where $\mathcal{L}$ represents the empirical risk, $\hat{Y}$ is the predicted output, and $\lambda_{reg}$ denotes the regularization strength. $\mathcal{R}(\hat{\bm{\beta_c}}, \hat{\bm{\beta_u}}, \hat{\bm{\beta_s}})$ represents the regularization term trained parameters. Bold symbols denote vectors or matrices, and symbols with $\hat{}$ indicate predicted values, while those without denote ground truth.
}

In this paper, we analyze five regularization methods for mitigating shortcut learning: Lasso (L1) regularization, Ridge (L2) regularization, causal regularization \citep{bahadori2017causal}, Expert Yielded Estimates (EYE) regularization \citep{wang2018learning,wang2022learning}, and causal effect regularization \citep{kumar2024causal}. Through both theoretical and empirical analyses, we evaluate these methods to assess their effectiveness under different conditions.

\textbf{L1 regularization.} L1 (lasso) regularization is commonly used to achieve sparse solutions \citep{vidaurre2013survey}. We apply L1 regularization solely to the parameters $\bm{\hat{\beta_{us}}}$, as penalizing $\bm{\hat{\beta_c}}$ (known concepts) is unnecessary, given that these concepts are established to causally influence the output.
\begin{align}
    \mathcal{R}_{L1}(\hat{\bm{\beta_c}}, \hat{\bm{\beta_u}}, \hat{\bm{\beta_s}})=||\bm{\hat{\beta_{us}}}||_1
\end{align}
where $p$-norm is defined as $$||\bm{x}||_p=\left(\sum_{i=1}^n|x_i|^p\right)^{\frac{1}{p}}$$
\textbf{L2 regularization.} L2 (ridge) regularization \citep{hoerl1970ridgea} is commonly used to mitigate the bias of coefficients \citep{hoerl1970ridgeb}. Similarly, we applied L2 regularization only to the parameters $\bm{\beta_u}$ and $\bm{\beta_s}$.
\begin{align}
    \mathcal{R}_{L2}(\hat{\bm{\beta_c}}, \hat{\bm{\beta_u}}, \hat{\bm{\beta_s}})=||\bm{\hat{\beta_{us}}}||_2^2
\end{align}

\textbf{EYE regularization.} EYE regularization \citep{wang2018learning, wang2022learning} is a combination of L1 and L2 regularizations. It is proposed to improve model credibility, encouraging denser coefficients for known concepts while promoting sparsity for other features. In our setting, it is formulated as
\begin{multline}
    \mathcal{R}_{EYE}(\hat{\bm{\beta_c}}, \hat{\bm{\beta_u}}, \hat{\bm{\beta_s}})=||\bm{\hat{\beta_{us}}}||_1\\
    +\sqrt{||\bm{\hat{\beta_{us}}}||_1^2+||\bm{\hat{\beta_c}}||^2_2}
\end{multline}

\textbf{Causal \& Causal effect regularization.} Causal regularization~\citep{bahadori2017causal} and causal effect regularization~\citep{kumar2024causal} are variants of L2 regularization. The general form of these two regularization methods are
\begin{align}
\mathcal{R}_{caus}(\hat{\bm{\beta_c}}, \hat{\bm{\beta_u}}, \hat{\bm{\beta_s}}) = \mathcal{R}_{caus}(\bm{\hat{\beta}})=\sum_{i=1}^h\lambda_i||\bm{\hat{\beta}}_i||^2_2
\end{align}
where $\lambda_{i}$ is the hyperparameter assigned to the coefficient $\bm{\hat{\beta}}_i$. For causal regularization, 
\begin{align}
    \lambda_i=1/P(\text{$H_i$ is the cause of $Y$})
\end{align}
For causal effect regularization,
\begin{align}
    \lambda_i= 1/\text{Estimated treatment effect of }(H_i)
\end{align}
In this paper, we treat these two regularization methods as equivalent because both assign coefficients that reflect the causal properties of features to the L2 regularization terms. Consequently, we use causal effect regularization as the representative method.
%%%%%%%%%%%%%%%%%%%%%%%%%%%%%%%%%%%%%%%%%%%%%%%%%%%%%%%%%%%%%%%%%%%%%%%%%%%%
%%%%%%%%%%%%%%%%%%%%%%%%%%%%%%%%%%%%%%%%%%%%%%%%%%%%%%%%%%%%%%%%%%%%%%%%%%%%
%%%%%%%%%%%%%%%%%%%%%%%%%%%%%%%%%%%%%%%%%%%%%%%%%%%%%%%%%%%%%%%%%%%%%%%%%%%%
\vspace{-.1cm}\section{Theoretical Analysis of Shortcut Mitigation Methods}\label{section:theory}
% \subsection{Shortcut mitigation}
\vspace{-0.3em}We first analyze the effectiveness of these regularization terms in shortcut mitigation. We conclude that L2, causal and causal effect regularization methods guarantee to alleviate the shortcut effects, i.e. shrink the absolute values of $\bm{\hat{\beta_s}}$ while the two other regularization methods cannot. The statements is formally structured as Proposition 1.

\textbf{Proposition 1}\label{proposition1}: \textit{Follow the problem setting~\eqref{problem_setting}, denote the trained parameters with regularization by $$\bm{\hat{\beta}}^{(\lambda)}=[\bm{\hat{\beta_c}}^{(\lambda)}\hspace{0.5em}\bm{\hat{\beta_u}}^{(\lambda)}\hspace{0.5em}\bm{\hat{\beta_s}}^{(\lambda)}]$$
where $\lambda$ is the regularization strength. When $\lambda=0$, the model is trained without regularization.
(i)Training with L1 and EYE does not guarantee to alleviate the shortcut effects.
(ii) Training with L2, causal and causal effect regularization can help mitigate shortcut to some extent, which is
$$|\hat{\bm{\beta}_s}^{reg}|\leq|\hat{\bm{\beta}_s}^{without \_reg}|$$}

\textbf{Proof sketch}. (i) For L1 regularization, consider the case where $U$ and $S$ are perfectly correlated. Because L1 regularization will generate sparse results, the results can be either $\hat{\beta_s}=\beta_u$ and $\hat{\beta_u}=0$ or vice versa. When the result is $\hat{\beta_s}=\beta_u$ and $\hat{\beta_u}=0$, the shortcut effects are exacerbated rather than mitigated. 

For EYE regularization, the coupling penalty on $\sqrt{||\bm{\hat{\beta_{us}}}||_1^2+||\bm{\hat{\beta_c}}||^2_2}$ creates a trade-off between $\bm{\hat{\beta_{us}}}$ and $\bm{\hat{\beta_c}}$. Hence, decreasing $\bm{\hat{\beta_c}}$ to reduce the penalty may increase the need for larger $\bm{\hat{\beta_{us}}}$ to maintain the fit, which makes it possible to have $|\hat{\bm{\beta}_s}^{reg}|>|\hat{\bm{\beta}_s}^{without \_reg}|$. 

(ii) The loss function with L2, causal and causal effect regularization can be formulated as
\begin{equation}
    loss = \frac{1}{n}||\bm{Y}-\hat{\bm{Y}}||_2^2+\lambda||\bm{\mathcal{D}}_\lambda\bm{\hat{\beta}}_{us}||_2^2
\end{equation}
where $\mathcal{D}_\lambda$ is a diagonal matrix where the diagonal entities are the square root of the weights assigned to each coefficients in causal and causal effect regularization. For L2 regularization, $\mathcal{D}_\lambda=\mathbb{I}$ is an identity matrix.

Take derivative with respect to $\bm{\hat{\beta}_c}$ and $\bm{\hat{\beta_{us}}}$,
\begin{multline}
       \bm{\hat{\beta_{us}}}^{(\lambda)}=\frac{1}{n}\left(\frac{1}{n}\bm{H}_{us}^T\left(\bm{\mathbb{I}}_c-\bm{\Pi}_c\right)\bm{H}_{us}+\lambda\bm{\mathcal{D}}_\lambda^2\right)^{-1}\\
       \bm{H}_{us}^T\left(\bm{\mathbb{I}}_c-\bm{\Pi}_c\right)\bm{Y} 
\end{multline}
where $$\bm{H}_{us}=[\bm{U}\hspace{0.5em}\bm{S}],\hspace{0.5em}\bm{\Pi}_c = \bm{C}(\bm{C}^T\bm{C})^{-1}\bm{C}^T$$
Then $\bm{H}_{us}^T(\mathbb{I}-\bm{\Pi}_c)\bm{H}_{us}$ is symmetric and normal, thus can be decomposed as 
\begin{equation}
    \bm{H}_{us}^T(\mathbb{I}-\bm{\Pi}_c)\bm{H}_{us}=\bm{Q}\bm{\Lambda}\bm{Q}^T
\end{equation}
where $\bm{Q}$ is orthogonal matrix and $\bm{\Lambda}$ is a diagonal matrix with non-negative eigenvalues $\Lambda_i$. Then 
\begin{align}
    \bm{H}_{us}^T(\mathbb{I}-\bm{\Pi}_c)\bm{Y}=\bm{Q}\bm{z}\\
     \hat{\bm{\beta}}_{us}=\frac{1}{n}\left(\frac{1}{n}\bm{Q}\bm{\Lambda}\bm{Q}^T+\lambda\bm{\mathcal{D}}_\lambda^2\right)^{-1}\bm{Q}\bm{z}
\end{align}

Then, for $i-$th entity,
\begin{equation}
    \hat{\bm{\beta}}_{us,i}=\dfrac{n}{\frac{\Lambda_i}{n}+n\lambda\bm{\mathcal{D}}_{ii}^2}z_i'
\end{equation}

Hence, increasing regularization strength $\lambda$ increasing the positive denominator, which makes the absolute value of $\bm{\hat{\beta}}_{us,i}$ decrease. $\square$.
% Hence $$\hat{\bm{\beta}}^{(\lambda)}_{us,i}=\dfrac{n}{\frac{\Lambda_i}{n}+n\lambda\bm{\mathcal{D}}_{ii}^2}z_i'$$
% then $$|\bm{\hat{\beta}}_{us,i}^{(\lambda)}|\leq|\bm{\hat{\beta}}_{us,i}^{(0)}|\hspace{2em}\square$$
The entire proof is shown in the supplementary \ref{proof:proposition1}.

\textbf{Correlations between unknown concepts and shortcuts affect the regularization performance}. The proof indicates that L1 and EYE regularization can sometimes amplify the influence of shortcut variables, particularly when these variables correlate with unknown concepts. This amplification occurs because the regularization process cannot distinguish between shortcuts and unknown concepts, making it challenging for the model to determine which features to prioritize. Furthermore, L1 regularization typically produces sparse models that focus on fewer features while maintaining predictive performance. However, when shortcuts are highly correlated with unknown concepts, especially when considering combinations of several unknown concepts, L1 regularization may disproportionately emphasize the shortcut features to achieve sparsity. This can exacerbate the problem of shortcut learning.

\textbf{Regularization can over-regularize causally relevant features}. The proof of Proposition 1 demonstrates that while L2-based regularization aims to mitigate the effects of shortcuts, it also diminishes the influence of unknown concepts, potentially degrading model performance. This suppression is influenced by the eigenvalues $\bm{\Lambda}$ of $\bm{H}_{us}^T(\mathbb{I}-\bm{\Pi}c)\bm{H}_{us}$ and the coefficients $\bm{\mathcal{D}}_{ii}$ assigned to each regularization term. For L2 regularization, $\bm{\Lambda}$ alone determines the resulting weights. For causal and causal effect regularization, accurate estimations of the probability of causal correlations and treatment effects are also critical. Inaccurate estimations can cause models to wrongly rely on shortcut variables, undermining model's robustness.

Additionally, we aim to investigate scenarios where regularization methods effectively eliminate shortcut learning. We detail instances where each regularization approach succeeds in shortcut mitigation, as outlined in Proposition 2.

\textbf{Proposition 2}\label{proposition2}: \textit{Follow the problem setting, when $c=u=s=1$, and $S$ is correlated with $Y$ by correlating with $C$ and $U$, i.e. $C$ and $U$ are not co-linear, $S=\delta_cC+\delta_uU$, the regularization methods can eliminate the shortcuts under the following situations. (Without loss of generality, assume $\beta_c$ and $\beta_u$ are non negative, other cases can be derived in the same way.)}

\textit{(i) For L1 regularization: $\frac{\delta_c+\delta_u-1}{\delta_c}\leq0$}

\textit{(ii) For L2 regularization: $\frac{\beta_c+\beta_c\delta_u^2-2\beta_u\delta_c\delta_u}{\delta_c^2+\delta_u^2+1}\geq\beta_c$}

\textit{(iii) For EYE regularization: $\frac{2\beta_c-2\frac{\delta_c}{\delta_u-1}\beta_u}{\left(\frac{\delta_c}{\delta_u-1}\right)^2+1}\geq\beta_c$}

\textit{(iv) For causal and causal effect regularization, denote the coefficients assigned to $C$, $U$ and $S$ as $\lambda_c$, $\lambda_u$ and $\lambda_s$, the condition is $\frac{\beta_c+\frac{\lambda_u}{\lambda_s}\delta_u^2\beta_c-\frac{\lambda_u}{\lambda_s}\delta_u\delta_c\beta_u}{\frac{\lambda_c}{\lambda_s}\delta_c^2+{\frac{\lambda_u}{\lambda_s}\delta_u^2+1}}=\beta_c$}

\textbf{Proof sketch}. Given the simplified linear relationship between input variables and output as
\begin{equation}
    Y_i=\beta_cC_i+\beta_uU_i+\beta_sS_i+\epsilon_i, i\in\{1,2,\cdots,n\}
\end{equation}
where the non-bold symbols indicate scalars rather than vectors. Choose the regularization strength $\lambda$ so that after regularization, the loss for $Y$ and $\hat{Y}$ is no less than without regularization. Further, for any $\hat{\beta_c}$, we can get the same empirical risk (empirical risk for noise terms) when $\hat{\beta}_u$ and $\hat{\beta}_s$ follow:
\begin{align}
        \hat{\beta}_u=\beta_u+\hat{\beta}_c\frac{\delta_u}{\delta_c}-\beta_c\frac{\delta_u}{\delta_c}\hspace{2em}
        \hat{\beta}_s=\frac{\beta_c-\hat{\beta}_c}{\delta_c}
\end{align}
where notations without $\hat{}$ represent the true weights and represent trained weights with $\hat{}$ . Then for different regularization methods, the regularization term can be converted into a function with respect to $\hat{\beta}_c$. Taking the derivative with respect to $\hat{\beta}_c$ shows us when can the regularization be optimal. $\square$ 

The complete proof are shown in supplementary \ref{proof:proposition2}. 

\textbf{Data normalization enhances the effectiveness of regularization methods.} According to the proposition, if all features, including the output feature $Y$, are standardized with zero mean and unit variance, then $\delta_c + \delta_u = 1$. Consequently, for $L1$ regularization, the optimal condition is satisfied. Moreover, for EYE regularization, the optimal condition can be simplified to: \begin{equation} 
\beta_c + \beta_u \geq \beta_c 
\end{equation} This condition is inherently met, assuming $\beta_c$ and $\beta_u$ are non-negative, with $\beta_u$ also being non-zero.

\textbf{Accurate estimation of causal properties is essential for effectively mitigating shortcuts.} Proposition 2 indicates that when the causal properties of shortcut features, especially the probability of a feature causing the output and its treatment effect, are accurately estimated to be close to zero, the shortcuts are successfully mitigated. This is because when $\lambda_s\rightarrow\infty$, the relationship in (iv) of Proposition 2 becomes $\beta_c=\beta_c$, which is inherently satisfied. Thus, precise estimation of causal properties is crucial for reducing shortcuts and uncovering the underlying causal structure of the data.
%%%%%%%%%%%%%%%%%%%%%%%%%%%%%%%%%%%%%%%%%%%%%%%%%%%%%%%%%%%%%%%%%%%%%%%%%%%%
%%%%%%%%%%%%%%%%%%%%%%%%%%%%%%%%%%%%%%%%%%%%%%%%%%%%%%%%%%%%%%%%%%%%%%%%%%%%
%%%%%%%%%%%%%%%%%%%%%%%%%%%%%%%%%%%%%%%%%%%%%%%%%%%%%%%%%%%%%%%%%%%%%%%%%%%%
\vspace{-.2cm}\section{Experiments}
In this section, we first explore the success and failure of regularization methods in mitigating shortcuts using synthetic datasets. Additionally, we demonstrate how these regularization methods work on three different real datasets to reduce shortcuts and the key factors that affect the performance. The results generate insights that align with our theoretical framework. The codes for experiments are available at \url{https://github.com/ai4ai-lab/regularization_for_shortcut_migtigation}.

\textbf{Baselines.} For each dataset, we initially train the model without any regularization to establish a baseline for using shortcut variables in prediction. Subsequently, we apply L1, L2, EYE, and causal effect regularization to assess their impact. We exclude causal regularization for two reasons. First, it performs similarly to causal effect regularization due to their analogous designs. Second, the dataset required to train the coefficients for each regularization term in causal regularization is not publicly available. Consequently, we use causal effect regularization as a representative method to assess performance.

\textbf{Synthetic data.} We generate three types of synthetic datasets (differing in terms of the relationship between shortcut variables and other variables) following:
\vspace{-0.5em}
\begin{enumerate}
\setlength\itemsep{-0.2em}
    \item Extract two concept variables ($\bm{C}=[C_1, C_2]$) and two unknown concepts ($\bm{U}=[U_1, U_2]$) from standard normal distribution $\mathcal{N}(0,1)$.
    \item Generate output feature ($Y$) based on a linear combination of $\bm{C}$ and $\bm{U}$ (independent of $S$).
    \item Generate one shortcut $S$ variable in three different ways:
    \vspace{-0.1em}
    \begin{itemize}
        \item Only correlating with $\bm{C}$
        \item Correlating with $\bm{U}$ and partial of $\bm{C}$
        \item Adding a random noise on $Y$
    \end{itemize}
    \vspace{-1em}
\end{enumerate}

\begin{figure*}[ht!]
    \centering
    \includegraphics[width=0.85\linewidth]{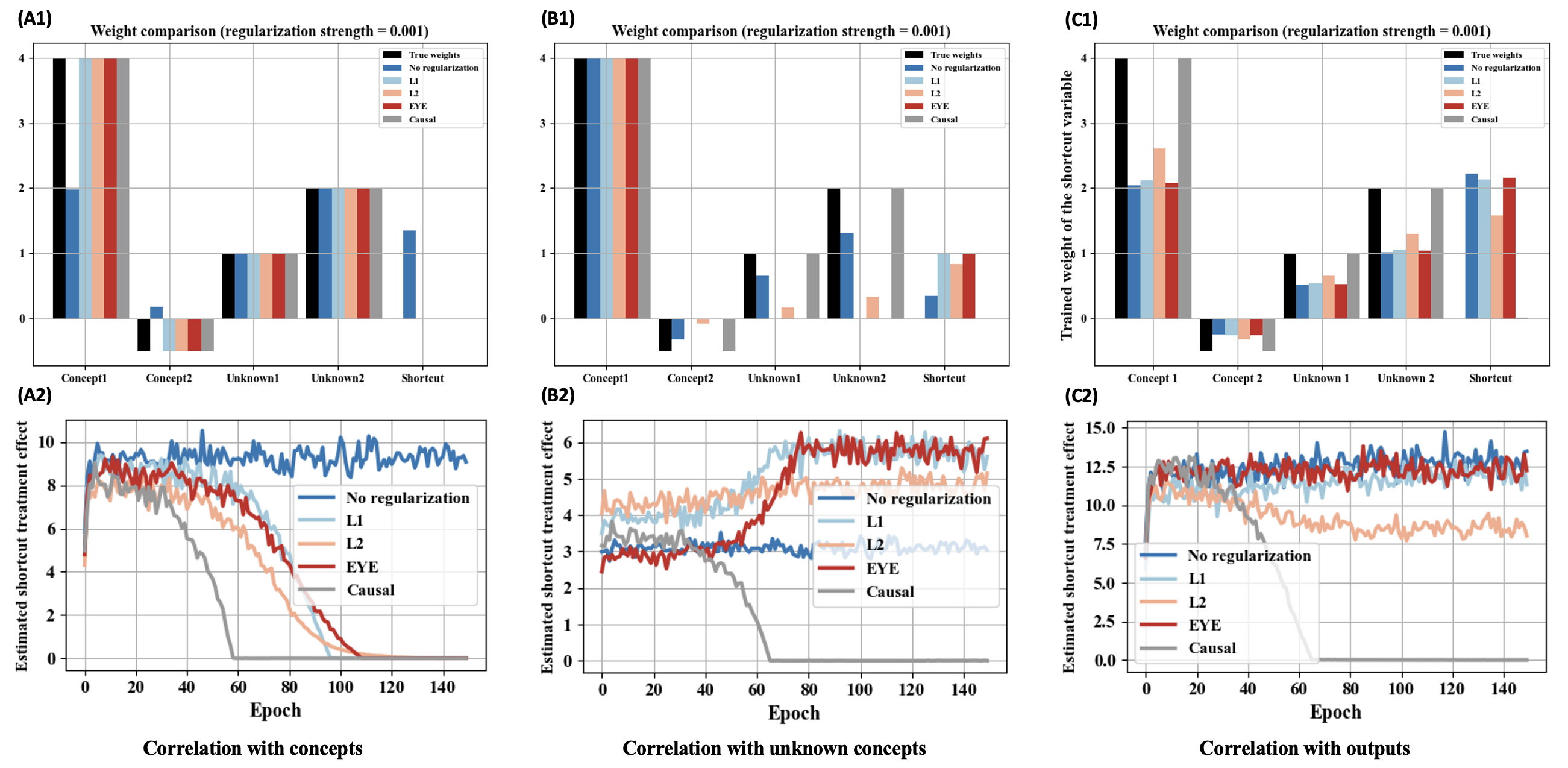}
    \caption{Comparison of trained weights and estimated treatment effects across training epochs. (A) Regularization are successful when $S$ is only correalted with $\bm{C}$. (B) Regularization fail when $S$ is correlated with $\bm{U}$. (C) Regularization fail when $S$ is highly correlated with $Y$. In the weight comparison plots, black bars represent the true weights assigned to each variable, consistent across different shortcut scenarios. The treatment effect plots depict the estimated treatment effect of the shortcut variable throughout the training epochs.}
    \label{fig:fail_case}
    \vspace{-1em}
\end{figure*}

\begin{figure*}
    \centering
    \includegraphics[width=0.85\linewidth]{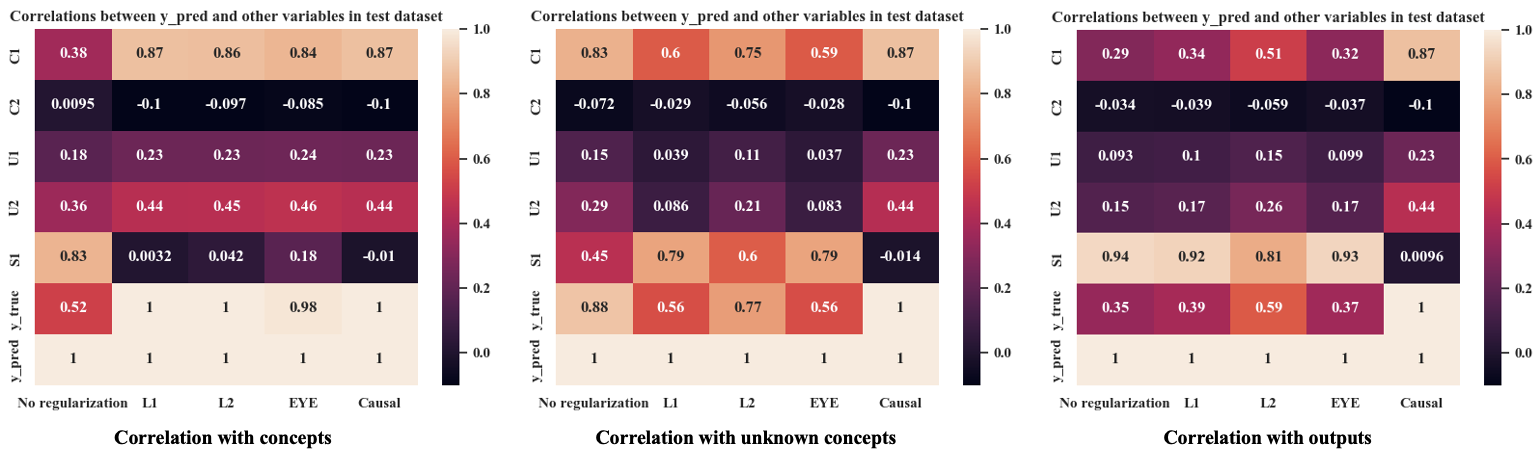}
    \caption{Correlation between predicted values and all variables in synthetic experiments, including true outputs in the test dataset, under different regularization methods. As the correlation between outputs and unknown concepts increases, regularization methods become less effective in mitigating shortcut dependencies.}
    \label{fig:y_pred_correlation}
\end{figure*}

For each shortcut variable, correlations with input and output features exist only in the training data. In the test set, $S$ is sampled from a standard normal distribution. For the synthetic dataset, we use a linear model with standardized inputs and apply each regularization method with a strength of $0.001$. For causal effect regularization, the treatment effect of the shortcut is set to $0.001$, while it is set to $1$ for the other variables (details regarding the experimental settings can be found in supplementary \ref{synthetic_dataset}).

\vspace{-0.5em}\subsection{Results on Synthetic Data}
\vspace{-0.5em}To evaluate shortcut mitigation, we employ two metrics: trained weights and the estimated treatment effect of $S$ across epochs. Using a synthetic dataset with known ground truth enables comparison between true and regularized weights, illustrating the effectiveness of regularization in reducing shortcut reliance.

To estimate the treatment effect, we adapt the T-learner approach. Since $S$ is continuous, we randomly sample 100 instances from a standard normal distribution, input them into the model, and compute the standard deviation of the outputs as the estimated treatment effect. Effective mitigation should result in uniform outputs with a standard deviation of zero. The results are shown in Figure \ref{fig:fail_case}.

\textbf{Regularization methods succeed when shortcuts correlate solely with known concepts.} As shown in Figure \ref{fig:fail_case}, panel A demonstrates that when $S$ correlates only with $\bm{C}$, all regularization methods perform effectively. This is because $\bm{C}$, which are not penalized by L1 and L2 regularization, contain all relevant information from $S$. Consequently, the model shifts its predictive reliance towards $\bm{U}$, reducing dependence on shortcuts. This prioritization allows regularization to effectively mitigate shortcut reliance. Additionally, the estimated treatment effects plot shows that causal effect regularization converges fastet than the other methods, highlighting its advantage.

Furthermore, we investigated the mitigation performance in relation to the relationship between the shortcut variable $S$ and the unknown concepts $\bm{U}$. We conducted experiments on various correlations between $S$ and $\bm{U}$, ranging from $0$ to $0.5$. The results indicate that as the correlation between $S$ and $\bm{U}$ increases, the mitigation performance of the regularization methods declines rapidly. This highlights a limitation of regularization methods when the shortcut variable is highly correlated with unknown concepts. The detailed results are presented in Appendix \ref{correlation_with_unknown_concepts}.

\textbf{Regularization methods fail when shortcuts correlate with unknown concepts.} Panel B shows that when $S$ is partially correlated with $\bm{U}$, all regularization methods fail, except for causal effect regularization with accurately estimated treatment effects. The failure of the other methods aligns with the findings from Proposition 1. Specifically, L1 exacerbates shortcut learning by allocating all weights to $S$, favoring sparsity by prioritizing one $S$ that provides similar predictive information as using the two unknown concepts. L2, on the other hand, distributes weights among both $\bm{U}$ and $S$, also fails to effectively mitigate shortcut reliance. The treatment effect plots reveal that the application of L1, L2, and EYE unexpectedly increases the model's dependence on $S$, which is contrary to the desired outcome.

\textbf{Regularization methods fail when shortcuts are highly correlated with the output.} Panel C reveals that when $S$ is derived directly from the $Y$, specifically by adding random noise, all regularization methods except for causal effect regularization fail.  In this case, the correlation between $\bm{U}$, $\bm{C}$ and $S$ is the strongest among three scenarios, which might be the reason to fail. Further, the treatment effect plot underscores this outcome, showing that the regularization approaches, aside from causal effect, fail to reduce the model's reliance on the shortcut variable, learning nothing to diminish this dependency during training.

Figure \ref{fig:y_pred_correlation} illustrates the correlation between different regularization methods and predicted values. The results indicate that only when the shortcut variable is correlated with concepts do regularization methods effectively reduce its correlation with predictions to near zero. Otherwise, they provide only limited mitigation.

\textbf{Complex model structures diminish the effectiveness of regularization methods.} Furthermore, relaxing the assumption of a linear relationship (Assumption \ref{assumption2}) allows us to apply regularization methods to neural networks. For instance, in a multi-layer network with a first layer consisting of 10 neurons and 5 input variables, the parameter matrix has a shape of $(10, 5)$. We can apply regularization to these parameters, designating the first two columns for known concepts ($\beta_c$) and the next two for unknown concepts ($\beta_u$). We conducted this experiment to investigate the application of regularization in non-linear models. The results of this experiment are presented in Figure \ref{fig:te_neural_network}. In comparison to Figure \ref{fig:fail_case}, when the shortcut is only correlated with known concepts, regularization methods are effective in linear models but not in neural networks. This discrepancy may be attributed to the deep architectures of neural networks, which can diminish the impact of regularization in mitigating shortcuts. Therefore, when applying regularization to address shortcuts, it is advisable to utilize simpler models and avoid overly complex structures. More details are shown in Appendix \ref{nonlinear_case}.

\begin{figure*}[ht]
    \centering
    \includegraphics[width=\linewidth]{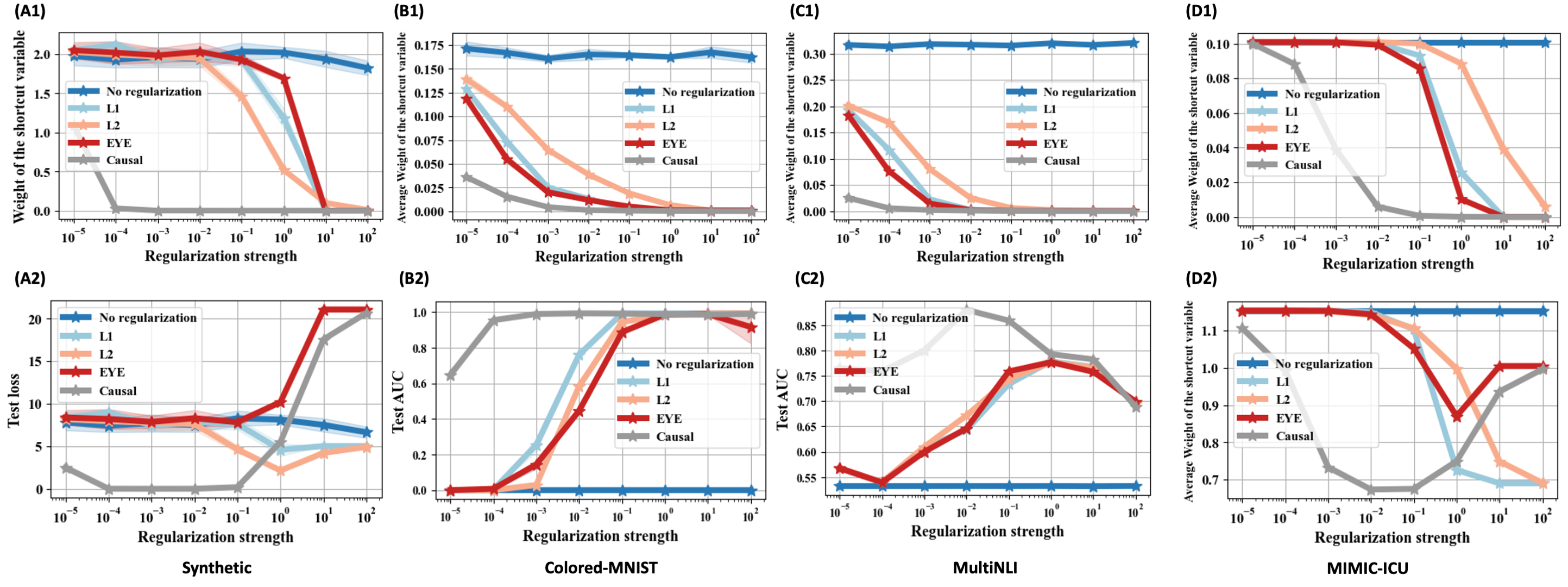}
    \caption{Comparison of weights and test loss (AUC for classification problems and MSE for regression problem) across varying regularization strengths for synthetic dataset, Colored-MNIST, MultiNLI, and MIMIC-IV dataset. Experiments were conducted ten times; the solid line represents the mean results, while the shaded area indicates the standard error of the ten experiments.}
    \label{fig:Results}
\end{figure*}

% \begin{figure*}[ht]
%     \centering
%     \includegraphics[width=0.8\linewidth]{image/correlation_verification.png}
%     \caption{Correlation among the data for synthetic, Colored-MNIST and MultiNLI datasets.}
%     \label{fig:correlation_verification}
%     \vspace{-0.5em}
% \end{figure*}

% SAME FIGURE AS ABOVE WITH LONGER CAPTION
% \begin{figure*}[ht]
%     \centering
%     \includegraphics[width=0.65\linewidth]{image/correlation_verification.png}
%     \caption{Correlation matrices across three datasets (Synthetic, Colored-MNIST, MultiNLI). In the Synthetic dataset, perfect correlations between shortcuts, concepts, and labels prevent regularization success. In Colored-MNIST, strong shortcut-concept correlations enable effective regularization, while MultiNLI's partial and complex correlations lead to incomplete mitigation.}
%     \label{fig:correlation_verification}
%     \vspace{-0.5em}
% \end{figure*}

In addition to the synthetic dataset, we explore the impact of regularization methods on real datasets and how varying the regularization strength affects model performance. We selected three real-world datasets, Colored-MNIST, MultiNLI and MIMIC-ICU for this analysis. We extracted concepts $\bm{C}$, unknown concepts $\bm{U}$, and shortcuts $\bm{S}$ from these datasets to demonstrate that the performance of the regularization methods align with the previous findings.

\vspace{-0.5em}\subsection{Results on Real Data.}
\vspace{-0.5em}\textbf{Colored-MNIST}. 
Following \cite{arjovsky2019invariant}, we adapted the \texttt{MNIST} dataset \citep{lecun1998gradient} by (i) binarizing the labels (digits 0-4 as one class, 5-9 as the other), and (ii) setting color labels $s$ equal to $y$ in the training set while reversing this in the test set. The image color (green or red) serves as a shortcut for predicting the binary label. We also created an unbiased dataset with random color assignments. Two models were trained: one for binary label prediction and one for color prediction, extracting concepts from the former and shortcuts from the latter. For unknown concept extraction, we used a pre-trained \texttt{resnet50} from \texttt{torchvision}.

% Following \cite{arjovsky2019invariant}, we adapted the \texttt{MNIST} dataset \citep{lecun1998gradient} for our experiments by (i) binarizing the labels into two classes, with digits 0-4 representing one class and 5-9 the other, (ii) setting color labels $s$ equal to $y$ in the training dataset and reversing this correlation in the test dataset. The color of each image (either green or red) acts as a potential shortcut for predicting the binary label. Additionally, we created an unbiased dataset where the color of each image is assigned randomly. Using this dataset, we trained two models: one to predict the binary label and another to predict the image color, utilizing the former to extract concepts and the latter to extract shortcuts. For the extraction of unknown concepts, we employed a pre-trained \texttt{resnet50} model from \texttt{torchvision}.

\textbf{MultiNLI}. 
For MultiNLI dataset \citep{williams2017broad}, the task is to classify whether a hypothesis is entailed or contradicts a premise. Previous work found a spurious correlation between contradictions and negation words \citep{gururangan2018annotation}. We focus on the binary task of distinguishing entailment from contradiction, introducing shortcuts by ensuring contradiction samples in the training set contain negation words. In the test set, entailment samples include negation words, breaking the correlation. We also created an unbiased dataset where negation and contradictions are independent. Two models were trained to extract concepts (true labels) and shortcuts (negation words), with unknown concept extraction using a pre-trained \texttt{BERT} model \citep{kenton2019bert}. We trained models without regularization, then applied L1, L2, EYE, and causal effect regularization. Shortcut mitigation was evaluated using two metrics: (1) average absolute weights of shortcut variables, with values close to zero indicating reduced shortcut reliance, and (2) AUC on the test set.
% For the MultiNLI dataset \citep{williams2017broad}, the task is to classify whether a hypothesis is entailed by or contradicts a given premise. Previous studies have identified a spurious correlation between contradictions and the presence of negation words \citep{gururangan2018annotation}. In our experiment, we focus on the binary task, distinguishing between entailment and contradiction. To introduce shortcuts, we curated a training dataset where contradiction samples contain negation words. In the test dataset, entailment samples contain negation words, breaking this correlation. As with Colored-MNIST, we also created an unbiased dataset where negation words and contradictions are independent. We trained two models to extract concepts (true labels) and shortcuts (negation words). For unknown concept extraction, we used a pre-trained \texttt{BERT} model \citep{kenton2019bert}.

% We first trained the model without regularization. Then, we applied L1, L2, EYE, and causal effect regularization. To evaluate the effectiveness of shortcut mitigation, we used two metrics: (1) the average absolute weights of the shortcut variables, where values close to zero indicate reduced reliance on shortcuts, and (2) the AUC on the test dataset. Since the relationship between shortcuts and outputs is reversed in the test dataset, strong performance on both training and test sets indicates successful mitigation of shortcut. To determine the optimal regularization strength, we conducted experiments across varying strengths. The results are presented in Figure \ref{fig:Results}.

\textbf{MIMIC-ICU}. We conducted further experiments using MIMIC-IV database \citep{johnson2023mimic}, focusing on predicting the length of stay (LOS) in intensive care units (ICU). This regression problem differs from the previous classification experiments. We identified 19 variables for predicting LOS, categorizing them into known and unknown concepts based on their correlations with LOS. Specifically, variables with correlations greater than 0.1 with LOS are classified as known concepts, while the others are considered unknown. This categorization is necessary due to the lack of generally pre-trained models for extracting information from this dataset. Finally, we got 12 known concepts and 7 unknown concepts. We also generated 10 shortcut variables based on LOS, all of which are highly correlated with LOS in the training dataset but randomly distributed in the test dataset. As the other experiments, we applied all regularization methods to train a regression model, using mean square error (MSE) to evaluate model performance. The results on real-world datasets are shown in Figure \ref{fig:Results}. The detailed experimental settings are shown in the Appendix \ref{real_dataset}.

\textbf{Mitigation performance depends on the correlation}. 
Figure \ref{fig:Results} shows that, in the synthetic dataset, all regularization strengths fail to mitigate shortcuts, except for causal effect regularization. In the Colored-MNIST dataset, proper regularization effectively mitigates shortcuts. However, in the MultiNLI dataset, while regularization strengths above $0.01$ reduce shortcuts, they also harm predictive performance. This may be due to correlations within input features. As seen in Figure~\ref{fig:correlation_verification}, the synthetic dataset has perfect correlation between the shortcut and output, preventing regularization success. In MultiNLI, shortcuts are only partially correlated with unknown concepts, leading to incomplete mitigation. In contrast, Colored-MNIST has shortcut-concept correlations, allowing regularization to work. Further, in MIMIC-ICU, shortcuts are highly correlated with the output, making it difficult to mitigate. Thus, shortcut mitigation success depends on the correlations between input features.

% The plot reveals that, in the synthetic dataset, all regularization strengths fail to help mitigate shortcuts (except for causal effect regularization). In the Colored-MNIST dataset, selecting the proper regularization strength effectively eliminates shortcuts. However, in the MultiNLI dataset, although shortcuts are mitigated when the regularization strength exceeds $0.01$, this comes at the cost of predictive performance. This behavior may be attributed to correlations within the input features. As shown in the heatmap in Figure~\ref{fig:correlation_verification}, the synthetic dataset has a perfect correlation between the shortcut and the output, making it impossible for regularization methods to succeed. In the MultiNLI dataset, shortcut variables are only partially correlated with unknown concepts, leading to imperfect mitigation. In contrast, the Colored-MNIST dataset shows correlation only between shortcut variables and concepts, allowing regularization methods to perform well. Therefore, the success of shortcut mitigation is strongly influenced by the types of correlations among input features.
\textbf{Choice of regularization strength is critical}. The plots show that when the regularization strength is too small, all methods fail to mitigate the shortcuts. However, if the strength is too large, the model eliminates the shortcuts but loses its predictive ability. Additionally, the results also highlight the sensitivity of shortcut mitigation to regularization strength. For example, in the Colored-MNIST dataset, increasing the regularization strength from $10^{-3}$ to $10^{-1}$ led to an AUC improvement of nearly 0.9. Furthermore, althought the results indicate that the choice of regularization strength is critical, the optimal value for regularization strength can spread a large range. As shown in panal A of Figure \ref{fig:Results} (varying regularization strengths vs. model performance), there is a relatively large range of regularization strengths ($10^{-4}$ to $10^{-1}$) within which shortcut mitigation remains effective. This indicates a degree of robustness in the choice of regularization strength.
\vspace{-0.3cm} \section{Practical Considerations}
\vspace{-0.5em}When applying regularization techniques such as L1, L2 and EYE to mitigate shortcut learning, several key considerations can help practitioners avoid common pitfalls and achieve better results:

\textbf{Task-Specific Application.} Regularization should be adapted to the complexity of the task. In high-dimensional tasks with multi-modal data, it is important to balance limiting shortcut learning while maintaining model flexibility. Over-regularization risks underfitting, while under-regularization may fail to effectively reduce shortcut reliance. Regularization should be adjusted flexibly based on the feature space.

% \textbf{Regularization Strength Matters.} Regularization strength is critical but not a one-size-fits-all solution. Different strengths should be explored to find the right balance between reducing shortcut learning and maintaining task performance. Too little regularization may fail to address shortcut reliance, while too much could suppress important task-related information. Sweeping across various strengths helps to determine how sensitive the model is to changes in regularization. If performance varies significantly with different strengths, it may be advisable to explore alternative techniques, particularly in tasks with complex feature spaces.

% \textbf{Handling Uncertainty in OOD Scenarios and Data Collection.} One major challenge in shortcut learning is handling uncertainty when the model encounters OOD data. If shortcuts are not fully mitigated, the model may still rely on them, increasing uncertainty in these scenarios. Regularization alone may not suffice, especially with limited or imbalanced training data. Expanding and diversifying the training data, particularly with OOD samples, can expose the model to more feature interactions, reducing shortcut reliance and improving generalization. Testing regularization across various OOD scenarios is crucial to assess robustness and ensure the model generalizes beyond the training distribution.

\textbf{Regularization Strength Matters.} Regularization strength is critical but not a one-size-fits-all solution. Different strengths should be explored to find the right balance between reducing shortcut learning and maintaining task performance. Too little regularization may fail to address shortcut reliance, while too much could suppress important task-related information. Testing various strengths reveals the model's sensitivity to regularization. If performance is highly sensitive to this, alternative techniques may be worth considering, especially in complex tasks.

\textbf{Handling Uncertainty in OOD Scenarios and Data Collection.} Handling uncertainty in OOD data is a significant challenge in shortcut learning. If shortcuts are not fully mitigated, the model may still rely on them, increasing uncertainty. Regularization alone may not be enough, particularly with limited or imbalanced training data. Expanding and diversifying the dataset, especially with OOD samples, can reduce shortcut reliance and improve generalization. Testing regularization in multiple OOD scenarios helps ensure the model generalizes well beyond training data.

% One of the main challenges in shortcut learning is dealing with the uncertainty that arises when the model is tested on OOD data. If shortcuts are not fully mitigated, the model may still rely on them, increasing uncertainty in these settings. Regularization alone may not fully address this issue, especially if the training data is limited or imbalanced. Therefore, increasing the diversity and size of training data, particularly from OOD sources, can help expose the model to a wider range of feature interactions. This, in turn, can help reduce the shortcuts and strengthen generalization. Testing regularization techniques across multiple OOD scenarios becomes essential to evaluate their robustness and ensure that the model generalizes well beyond the training distribution.

\textbf{Refining Causal Concept Variables.} Regularization techniques often treat causal variables as static and perfectly defined, but real-world tasks are more dynamic. If shortcuts continue to influence the model after regularization, it may indicate that the initial set of causal variables is insufficient. Practitioners should be open to refining the set of concepts based on performance, especially in OOD contexts, as this can reduce reliance on shortcuts. Standard regularization methods such as L1, L2, EYE are agnostic to the treatment effect and constrain learning without knowledge of this process. However, the treatment effect is crucial to ensure the underlying causal process is captured. Alternatively, one might consider iteratively refining and updating assumptions about the causal variables to improve model robustness.

% \textbf{Conceptual Overlap Between Shortcuts and Causal Features and Complementary Approaches.} In practice, shortcuts and causal features can overlap, making disentangling these dependencies difficult. Regularization techniques alone may not fully distinguish between spurious correlations and causal relationships. While regularization can help reduce shortcut reliance, it is not a guaranteed solution when causal and shortcut features are closely intertwined. Techniques like causal inference or adversarial training could be combined with regularization to better address this challenge. Additionally, even causal regularization methods, such as those proposed by \cite{kumar2024causal}, which regularize the effect of attributes proportional to their average causal effect, struggle with assumptions about the data generation process. These assumptions may not hold in practice, or the required knowledge of the data process may not be available in advance. This suggests that relying solely on regularization methods—even those that are more explicitly tied to causality—can have limitations.

%%%%%%%%%%%%%%%%%%%%%%%%%%%%%%%%%%%%%%%%%%%%%%%%%%%%%%%%%%%%%%%%%%%%%%%%%%%%
%%%%%%%%%%%%%%%%%%%%%%%%%%%%%%%%%%%%%%%%%%%%%%%%%%%%%%%%%%%%%%%%%%%%%%%%%%%%
%%%%%%%%%%%%%%%%%%%%%%%%%%%%%%%%%%%%%%%%%%%%%%%%%%%%%%%%%%%%%%%%%%%%%%%%%%%%
\vspace{-0.3cm}\section{Conclusion}
\vspace{-0.6em}Regularization methods are helpful for mitigating shortcut learning but face challenges. Our analysis shows that techniques like L1, L2 and EYE often over-regularize, suppressing both spurious and causal features and limiting their effectiveness. Theoretical exploration and experiments on synthetic and real-world datasets indicate that current methods inadequately disentangle causal features from shortcuts. Future work should aim to develop techniques that better balance shortcut reduction with the preservation of causal information, enhancing model performance.

% \subsubsection*{Acknowledgements}
% All acknowledgments go at the end of the paper, including thanks to reviewers who gave useful comments, to colleagues who contributed to the ideas, and to funding agencies and corporate sponsors that provided financial support. 
% To preserve the anonymity, please include acknowledgments \emph{only} in the camera-ready papers.

\section*{Acknowledgements}
The authors would like to thank Jenna Wiens (University of Michigan) and Amit Sharma (Microsoft Research India) for valuable discussions regarding this work, as well as the anonymous
reviewers for constructive feedback.

\bibliography{ref}
% References follow the acknowledgements.  Use an unnumbered third level
% heading for the references section.  Please use the same font
% size for references as for the body of the paper---remember that
% references do not count against your page length total.

% \begin{thebibliography}{}
% \setlength{\itemindent}{-\leftmargin}
% \makeatletter\renewcommand{\@biblabel}[1]{}\makeatother
% \bibitem{} J.~Alspector, B.~Gupta, and R.~B.~Allen (1989).
%     \newblock Performance of a stochastic learning microchip.
%     \newblock In D. S. Touretzky (ed.),
%     \textit{Advances in Neural Information Processing Systems 1}, 748--760.
%     San Mateo, Calif.: Morgan Kaufmann.

% \bibitem{} F.~Rosenblatt (1962).
%     \newblock \textit{Principles of Neurodynamics.}
%     \newblock Washington, D.C.: Spartan Books.

% \bibitem{} G.~Tesauro (1989).
%     \newblock Neurogammon wins computer Olympiad.
%     \newblock \textit{Neural Computation} \textbf{1}(3):321--323.
% \end{thebibliography}

%%%%%%%%%%%%%%%%%%%%%%%%%%%%%%%%%%%%%%%%%%%%%%%%%%%%%%%%%%%%

\section*{Checklist}

% %%% BEGIN INSTRUCTIONS %%%

% %%% END INSTRUCTIONS %%%

\begin{enumerate}

\item For all models and algorithms presented, check if you include:
\begin{enumerate}
\item A clear description of the mathematical setting, assumptions, algorithm, and/or model. [Yes]
\item An analysis of the properties and complexity (time, space, sample size) of any algorithm. [Not Applicable]
\item (Optional) Anonymized source code, with specification of all dependencies, including external libraries. [Yes - will be made public after review process]
\end{enumerate}

\item For any theoretical claim, check if you include:
\begin{enumerate}
\item Statements of the full set of assumptions of all theoretical results. [Yes]
\item Complete proofs of all theoretical results. [Yes]
\item Clear explanations of any assumptions. [Yes]     
\end{enumerate}

\item For all figures and tables that present empirical results, check if you include:
\begin{enumerate}
\item The code, data, and instructions needed to reproduce the main experimental results (either in the supplemental material or as a URL). [Yes]
\item All the training details (e.g., data splits, hyperparameters, how they were chosen). [Yes]
        \item A clear definition of the specific measure or statistics and error bars (e.g., with respect to the random seed after running experiments multiple times). [Yes]
        \item A description of the computing infrastructure used. (e.g., type of GPUs, internal cluster, or cloud provider). [Not Applicable]
\end{enumerate}

\item If you are using existing assets (e.g., code, data, models) or curating/releasing new assets, check if you include:
\begin{enumerate}
\item Citations of the creator If your work uses existing assets. [Yes]
\item The license information of the assets, if applicable. [Not Applicable]
\item New assets either in the supplemental material or as a URL, if applicable. [Not Applicable]
\item Information about consent from data providers/curators. [Not Applicable]
\item Discussion of sensible content if applicable, e.g., personally identifiable information or offensive content. [Not Applicable]
\end{enumerate}

\item If you used crowdsourcing or conducted research with human subjects, check if you include:
\begin{enumerate}
\item The full text of instructions given to participants and screenshots. [Not Applicable]
\item Descriptions of potential participant risks, with links to Institutional Review Board (IRB) approvals if applicable. [Not Applicable]
\item The estimated hourly wage paid to participants and the total amount spent on participant compensation. [Not Applicable]
\end{enumerate}
\end{enumerate}

%%%%%%%%%%%%%%%%%%%%%%%%%%%%%%%%%%%%%%%%%%%%%%%%%%%%%%%%%%%%
\onecolumn
\aistatstitle{Supplementary Materials:\\
\vspace{0.5em} Do Regularization Methods for Shortcut Mitigation Work As Intended?}
\renewcommand{\thesection}{\Alph{section}}
\setcounter{section}{0}
\section{Proofs for section}

\subsection{Proof for Proposition 1}\label{proof:proposition1}

\textbf{Proposition 1}: \textit{Follow the problem setting~\eqref{problem_setting}, denote the trained parameters with regularization by $$\bm{\hat{\beta}}^{(\lambda)}=[\bm{\hat{\beta_c}}^{(\lambda)}\hspace{0.5em}\bm{\hat{\beta_u}}^{(\lambda)}\hspace{0.5em}\bm{\hat{\beta_s}}^{(\lambda)}]$$
where $\lambda$ is the regularization strength. When $\lambda=0$, the model is trained without regularization.}

\textit{(i) Training with L1 and EYE does not guarantee alleviation of the shortcut effects.}

\textit{(ii) Training with L2, causal and causal effect regularization can help mitigate shortcut to some extents, i.e.}
$$|\hat{\bm{\beta}_s}^{reg}|\leq|\hat{\bm{\beta}_s}^{without \_reg}|$$

\textbf{Proof}:

According to the problem setting~\eqref{problem_setting}, the assumed linear model is
\begin{equation}
    \hat{\bm{Y}} = \bm{C}\hat{\bm{\beta}}_c + \bm{U}\hat{\bm{\beta}}_u + \bm{S}\hat{\bm{\beta}}_s
\end{equation}

Loss used to train the model can be written as
\begin{equation}
    loss = \frac{1}{n}||\bm{Y}-\hat{\bm{Y}}||_2^2 + \lambda\mathcal{R}(\hat{\bm{\beta_c}}, \hat{\bm{\beta_u}}, \hat{\bm{\beta_s}})
\end{equation}
where $n$ is the number of samples in training dataset, $\lambda$ is the regularization strength and $\mathcal{R}(\hat{\bm{\beta_c}}, \hat{\bm{\beta_u}}, \hat{\bm{\beta_s}})$ is the regularization value according to different regularization methods. Further, the trained parameters are denoted as $\bm{\hat{\beta}}^{(\lambda)}=[\bm{\hat{\beta_c}}^{(\lambda)}\hspace{0.5em}\bm{\hat{\beta_u}}^{(\lambda)}\hspace{0.5em}\bm{\hat{\beta_s}}^{(\lambda)}]$. Some notations are listed as:
$$\bm{H}=[\bm{C}\hspace{0.5em}\bm{U}\hspace{0.5em}\bm{S}]\hspace{1em}\bm{H}_{us}=[\bm{U}\hspace{0.5em}\bm{S}]\hspace{1em}\bm{\hat{\beta}}_{us}=\begin{bmatrix}
    \bm{\hat{\beta}}_{u}\\
    \bm{\hat{\beta}}_{s}
\end{bmatrix}$$

\textbf{L1 regularization}

Prove via counterexample, suppose $\bm{\hat{\beta}}_c=0$ and $\bm{U}$ and $\bm{S}$ are perfectly correlated, then
\begin{equation}
    \hat{\bm{Y}} = \bm{C}\hat{\bm{\beta}}_c + \bm{U}\hat{\bm{\beta}}_u + \bm{S}\hat{\bm{\beta}}_s= \bm{U}(\hat{\bm{\beta}}_u + \hat{\bm{\beta}}_s)
\end{equation}

Then without regularization, the ordinary least square solution minimize
\begin{equation}
    \min_{\bm{\hat{\beta}}_u, \bm{\hat{\beta}}_s} \frac{1}{n}||\bm{Y}-\bm{U}(\hat{\bm{\beta}}_u + \hat{\bm{\beta}}_s)||_2^2
\end{equation}

Due to the perfect linear multicollinearity, there are infinitely many solutions for $\theta$ satisfying $\hat{\bm{\beta}}_u + \hat{\bm{\beta}}_s= \bm{\theta}$, where 
$\bm{\theta}$ is the OLS estimate of the combined coefficient:
\begin{equation}
    \bm{\theta} = \left(\bm{U}^T\bm{U}\right)^{-1}\bm{U}^T\bm{Y}
\end{equation}
A common choice in the unregularized case is to distribute the coefficient equally:
\begin{equation}
    \hat{\bm{\beta}}_u^{(0)} = \hat{\bm{\beta}}_s^{(0)} = \frac{\bm{\theta}}{2}
\end{equation}

With L1 regularization, the optimization problem becomes
\begin{equation}
        \min_{\bm{\hat{\beta}}_u, \bm{\hat{\beta}}_s} \frac{1}{n}||\bm{Y}-\bm{U}(\hat{\bm{\beta}}_u + \hat{\bm{\beta}}_s)||_2^2 + \lambda\left(||\bm{\hat{\beta}}_u||_1 + ||\bm{\hat{\beta}}_s||_1\right)
\end{equation}

Since $\bm{U}=\bm{S}$, the model depends only on $\theta=\bm{\hat{\beta}}_u+\bm{\hat{\beta}}_s$, but the penalty depends on $||\bm{\hat{\beta}}_u||_1+||\bm{\hat{\beta}}_s||_1$. To minimize the penalty while keeping $\bm{\theta}$ constant, the optimal strategy is to set one coefficient to zero and assign all the weight to the other. According to triangle inequality,
\begin{equation}
    ||\bm{\hat{\beta}}_u||_1 + ||\bm{\hat{\beta}}_s||_1 \geq ||\bm{\hat{\beta}}_u + \bm{\hat{\beta}}_s||_1 = ||\bm{\theta}||_1
\end{equation}

Hence, when $\bm{\theta}>0$:
\begin{equation}
    \bm{\hat{\beta}}_u^{(\lambda)}=\bm{\theta},\hspace{0.5em}\bm{\hat{\beta}}_s^{(\lambda)}=\bm{0}\hspace{0.5em}\text{OR}\hspace{0.5em}\bm{\hat{\beta}}_u^{(\lambda)}=\bm{0},\hspace{0.5em}\bm{\hat{\beta}}_s^{(\lambda)}=\bm{\theta}
\end{equation}
Hence, when $\bm{\hat{\beta}}_u^{(\lambda)}=\bm{0},\hspace{0.5em}\bm{\hat{\beta}}_s^{(\lambda)}=\bm{\theta}$, 
\begin{equation}
    ||\bm{\hat{\beta}}_s^{(\lambda)}||_1 = ||\bm{\theta}||_1 \geq ||\frac{\bm{\theta}}{2}||_1 = ||\bm{\hat{\beta}}_s^{(0)}||_1
\end{equation}

Therefore, L1 regularization does not reliably mitigate the shortcut term and, in some cases, may incorrectly assign all weights to the shortcut variables. $\square$

\textbf{EYE regularization}

For EYE regularization, the loss function is
\begin{align}
loss &= \frac{1}{n}||\bm{Y}-\hat{\bm{Y}}||_2^2+\lambda\left(||\bm{\hat{\beta_{us}}}||_1+\sqrt{||\bm{\hat{\beta_{us}}}||_1^2+||\bm{\hat{\beta_c}}||^2_2}\right)\\
    &= \frac{1}{n}||\bm{Y}-\bm{C}\bm{\hat{\beta}}_c-\bm{H}_{us}\hat{\bm{\beta}}_{us}||_2^2+\lambda\left(||\bm{\hat{\beta_{us}}}||_1+\sqrt{||\bm{\hat{\beta_{us}}}||_1^2+||\bm{\hat{\beta_c}}||^2_2}\right)
\end{align}
Take derivative with respect to $\bm{\hat{\beta}}_c$ and $\bm{\hat{\beta}}_{us}$,
\begin{align}
    \frac{\partial loss}{\partial \bm{\hat{\beta}}_c} =& -\frac{2}{n}\bm{C}^T\left(\bm{Y}-\bm{C}\bm{\hat{\beta}}_c-\bm{H}_{us}\hat{\bm{\beta}}_{us}\right) + \lambda\frac{\bm{\hat{\beta}}_c}{\sqrt{||\bm{\hat{\beta_{us}}}||_1^2+||\bm{\hat{\beta_c}}||^2_2}}\\
    \frac{\partial loss}{\partial \bm{\hat{\beta}}_{us}} =& -\frac{2}{n}\bm{H}_{us}^T\left(\bm{Y}-\bm{C}\bm{\hat{\beta}}_c-\bm{H}_{us}\hat{\bm{\beta}}_{us}\right) + \lambda\left(sign(\bm{\hat{\beta}}_{us})+\frac{sign(\bm{\hat{\beta}}_{us})||\bm{\hat{\beta}}_{us}||_1}{\sqrt{||\bm{\hat{\beta_{us}}}||_1^2+||\bm{\hat{\beta_c}}||^2_2}}\right)
\end{align}

Then, at the optimum, the gradients are zero. Then,
\begin{align}
    & \frac{2}{n}\bm{C}^T\left(\bm{Y}-\bm{C}\bm{\hat{\beta}}_c-\bm{H}_{us}\hat{\bm{\beta}}_{us}\right) = \lambda\frac{\bm{\hat{\beta}}_c}{\sqrt{||\bm{\hat{\beta_{us}}}||_1^2+||\bm{\hat{\beta_c}}||^2_2}}\\
    & \frac{2}{n}\bm{H}_{us}^T\left(\bm{Y}-\bm{C}\bm{\hat{\beta}}_c-\bm{H}_{us}\hat{\bm{\beta}}_{us}\right) = \lambda\left(sign(\bm{\hat{\beta}}_{us})+\frac{sign(\bm{\hat{\beta}}_{us})||\bm{\hat{\beta}}_{us}||_1}{\sqrt{||\bm{\hat{\beta_{us}}}||_1^2+||\bm{\hat{\beta_c}}||^2_2}}\right)
\end{align}

Suppose the inequality $|\hat{\bm{\beta}_s}^{reg}|\leq|\hat{\bm{\beta}_s}^{without \_reg}|$ always hold. Then to reduce the penalty, $\bm{\hat{\beta}}_c$ may be decreased, which leads to decreasing in $\sqrt{||\bm{\hat{\beta_{us}}}||_1^2+||\bm{\hat{\beta_c}}||^2_2}$. Then $\frac{||\bm{\hat{\beta}}_{us}||_1}{\sqrt{||\bm{\hat{\beta_{us}}}||_1^2+||\bm{\hat{\beta_c}}||^2_2}}$ increases. According to the optimality condition for $\bm{\hat{\beta}}_{uc}$, $(\bm{Y}-\bm{C}\bm{\hat{\beta}}_c-\bm{H}_{us}\hat{\bm{\beta}}_{us})$ must also increase, which may require increasing $||\bm{\hat{\beta}}_{us}||_1$. Therefore, the inequality does not always hold. $\square$

\textbf{L2, causal and causal effect regularization}

The loss function with L2, causal and causal effect regularization can be formulated as
\begin{align}
    loss &= \frac{1}{n}||\bm{Y}-\hat{\bm{Y}}||_2^2+\lambda||\bm{\mathcal{D}}_\lambda\bm{\hat{\beta}}_{us}||_2^2\\
    &= \frac{1}{n}||\bm{Y}-\bm{C}\bm{\hat{\beta}}_c-\bm{H}_{us}\hat{\bm{\beta}}_{us}||_2^2+\lambda||\bm{\mathcal{D}}_\lambda\bm{\hat{\beta}}_{us}||_2^2
\end{align}
where $\bm{\mathcal{D}}_\lambda$ is a diagonal matrix whose diagonal entities are
\begin{itemize}
    \item $1$ for L2 regularization
    \item $1/P(\text{$\bm{H}_{us, i}$ is the cause of $\bm{Y}$})$ for causal regularization
    \item $1/|\text{Estimated treatment effect of $\bm{H}_{us, i}$}|$ for causal effect regularization
\end{itemize}
Hence the diagonal entities of $\bm{\mathcal{D}}_\lambda$ are all positive. Take derivative with respect to $\bm{\hat{\beta}}_c$ and $\bm{\hat{\beta}}_{us}$,
\begin{align}
    \frac{\partial loss}{\partial \bm{\hat{\beta}}_c} =& -\frac{2}{n}\bm{C}^T\left(\bm{Y}-\bm{C}\bm{\hat{\beta}}_c-\bm{H}_{us}\hat{\bm{\beta}}_{us}\right)\\
    =& -\frac{2}{n}\left(\bm{C}^T\bm{Y}-\bm{C}^T\bm{C}\bm{\hat{\beta}}_c-\bm{C}^T\bm{H}_{us}\hat{\bm{\beta}}_{us}\right)\\
    \frac{\partial loss}{\partial \bm{\hat{\beta}}_{us}} =& -\frac{2}{n}\bm{H}_{us}^T\left(\bm{Y}-\bm{C}\bm{\hat{\beta}}_c-\bm{H}_{us}\hat{\bm{\beta}}_{us}\right) + 2\lambda\bm{\mathcal{D}}_\lambda^2\bm{\hat{\beta}}_{us}\\
    =&-\frac{2}{n}\left(\bm{H}_{us}^T\bm{Y}-\bm{H}_{us}^T\bm{C}\bm{\hat{\beta}}_c-\bm{H}_{us}^T\bm{H}_{us}\hat{\bm{\beta}}_{us}\right) + 2\lambda\bm{\mathcal{D}}_\lambda^2\bm{\hat{\beta}}_{us}
\end{align}
According to Karush-Kuhn-Tucker (KKT) condition, set the derivatives to $0$, then
\begin{align}
     &\frac{\partial loss}{\partial \bm{\hat{\beta}}_c} = 0 \Rightarrow  \bm{C}^T\bm{Y}-\bm{C}^T\bm{C}\bm{\hat{\beta}}_c-\bm{C}^T\bm{H}_{us}\hat{\bm{\beta}}_{us} = 0\\
    \Rightarrow& \bm{\hat{\beta}}_c = \left(\bm{C}^T\bm{C}\right)^{-1}\left(\bm{C}^T\bm{Y}-\bm{C}^T\bm{H}_{us}\hat{\bm{\beta}}_{us}\right)\\ &\hspace{1.1em}=\left(\bm{C}^T\bm{C}\right)^{-1}\bm{C}^T\bm{Y}-\left(\bm{C}^T\bm{C}\right)^{-1}\bm{C}^T\bm{H}_{us}\hat{\bm{\beta}}_{us}
\end{align}
\begin{align}
     &\frac{\partial loss}{\partial \bm{\hat{\beta}}_{us}} = 0 \Rightarrow \left(\frac{1}{n}\bm{H}_{us}^T\bm{H}_{us}+\lambda\bm{\mathcal{D}}_\lambda^2\right)\hat{\bm{\beta}}_{us}=\frac{1}{n}\bm{H}_{us}^T\bm{Y}-\frac{1}{n}\bm{H}_{us}^T\bm{C}\bm{\hat{\beta}}_c\\
     \Rightarrow & \left(\frac{1}{n}\bm{H}_{us}^T\bm{H}_{us}+\lambda\bm{\mathcal{D}}_\lambda^2\right)\hat{\bm{\beta}}_{us}=\frac{1}{n}\bm{H}_{us}^T\bm{Y}-\frac{1}{n}\bm{H}_{us}^T\bm{C}\left(\left(\bm{C}^T\bm{C}\right)^{-1}\bm{C}^T\bm{Y}-\left(\bm{C}^T\bm{C}\right)^{-1}\bm{C}^T\bm{H}_{us}\hat{\bm{\beta}}_{us}\right)\\
      \Rightarrow & \left(\frac{1}{n}\bm{H}_{us}^T\bm{H}_{us}+\lambda\bm{\mathcal{D}}_\lambda^2\right)\hat{\bm{\beta}}_{us}=\frac{1}{n}\bm{H}_{us}^T\bm{Y}-\frac{1}{n}\bm{H}_{us}^T\bm{C}\left(\bm{C}^T\bm{C}\right)^{-1}\bm{C}^T\bm{Y}+\frac{1}{n}\bm{H}_{us}^T\bm{C}\left(\bm{C}^T\bm{C}\right)^{-1}\bm{C}^T\bm{H}_{us}\hat{\bm{\beta}}_{us}
\end{align}
Denote $\bm{\Pi}_c=\bm{C}\left(\bm{C}^T\bm{C}\right)^{-1}\bm{C}^T$, then
\begin{align}
    \Rightarrow & \left(\frac{1}{n}\bm{H}_{us}^T\bm{H}_{us}+\lambda\bm{\mathcal{D}}_\lambda^2\right)\hat{\bm{\beta}}_{us}=\frac{1}{n}\bm{H}_{us}^T\bm{Y}-\frac{1}{n}\bm{H}_{us}^T\bm{\Pi}_c\bm{Y}+\frac{1}{n}\bm{H}_{us}^T\bm{\Pi}_c\bm{H}_{us}\hat{\bm{\beta}}_{us}\\
    \Rightarrow & \left(\frac{1}{n}\bm{H}_{us}^T\bm{H}_{us}+\lambda\bm{\mathcal{D}}_\lambda^2-\frac{1}{n}\bm{H}_{us}^T\bm{\Pi}_c\bm{H}_{us}\right)\hat{\bm{\beta}}_{us}=\frac{1}{n}\bm{H}_{us}^T\bm{Y}-\frac{1}{n}\bm{H}_{us}^T\bm{\Pi}_c\bm{Y}\\
    \Rightarrow & \left(\frac{1}{n}\bm{H}_{us}^T\left(\bm{\mathbb{I}}_c-\bm{\Pi}_c\right)\bm{H}_{us}+\lambda\bm{\mathcal{D}}_\lambda^2\right)\hat{\bm{\beta}}_{us}=\frac{1}{n}\bm{H}_{us}^T\left(\bm{\mathbb{I}}_c-\bm{\Pi}_c\right)\bm{Y}\\
    \Rightarrow & \hspace{0.25em}\hat{\bm{\beta}}_{us}=\frac{1}{n}\left(\frac{1}{n}\bm{H}_{us}^T\left(\bm{\mathbb{I}}_c-\bm{\Pi}_c\right)\bm{H}_{us}+\lambda\bm{\mathcal{D}}_\lambda^2\right)^{-1}\bm{H}_{us}^T\left(\bm{\mathbb{I}}_c-\bm{\Pi}_c\right)\bm{Y}
\end{align}

Because $\bm{H}_{us}^T\left(\bm{\mathbb{I}}_c-\bm{\Pi}_c\right)\bm{H}_{us}$ is symmetric and normal, it can be diagnolized as
\begin{equation}
    \bm{H}_{us}^T\left(\bm{\mathbb{I}}_c-\bm{\Pi}_c\right)\bm{H}_{us} = \bm{Q}\bm{\Lambda}\bm{Q}^T
\end{equation}

where $\bm{Q}$ is an orthonormal basis of $\mathbb{R}^{u+s}$, hence $\bm{H}_{us}^T\left(\bm{\mathbb{I}}_c-\bm{\Pi}_c\right)\bm{Y}$ can be written as a linear combination of $\bm{Q}$, i.e.
\begin{equation}
    \exists\hspace{0.5em}\bm{z}\text{ such that }\bm{H}_{us}^T\left(\bm{\mathbb{I}}_c-\bm{\Pi}_c\right)\bm{Y}=\bm{Q}\bm{z}
\end{equation}
Then,
\begin{equation}
    \hat{\bm{\beta}}_{us}=\frac{1}{n}\left(\frac{1}{n}\bm{Q}\bm{\Lambda}\bm{Q}^T+\lambda\bm{\mathcal{D}}_\lambda^2\right)^{-1}\bm{Q}\bm{z}
\end{equation}

Because $\bm{Q}$ is orthonormal basis, let $\bm{z}'=\bm{Q}^T\bm{z}$, we can focus on the part
\begin{equation}
    \frac{1}{n}\left(\frac{1}{n}\bm{\Lambda}+\lambda\bm{\mathcal{D}}_\lambda^2\right)^{-1}\bm{z}'
\end{equation}

Then, for $i-$th entity,
\begin{equation}
    \hat{\bm{\beta}}_{us,i}=\dfrac{n}{\frac{\Lambda_i}{n}+n\lambda\bm{\mathcal{D}}_{ii}^2}z_i'
\end{equation}

Hence, increasing regularization strength $\lambda$ increasing the positive denominator, which makes the absolute value of $\bm{\hat{\beta}}_{us,i}$ decrease. $\square$.

\subsection{Proof for Proposition 2}\label{proof:proposition2}

\textbf{Proposition 2}: \textit{Follow the problem setting, when $c=u=s=1$, and $S$ is correlated with $Y$ by correlating with $C$ and $U$, i.e. $C$ and $U$ are not co-linear, $S=\delta_cC+\delta_uU$, the regularization methods can eliminate the shortcuts under the following situations.}

\textit{(i) For L1 regularization: $\frac{\delta_c+\delta_u-1}{\delta_c}\leq0$}

\textit{(ii) For L2 regularization: $\frac{\beta_c+\beta_c\delta_u^2-2\beta_u\delta_c\delta_u}{\delta_c^2+\delta_u^2+1}\geq\beta_c$}

\textit{(iii) For EYE regularization: $\frac{2\beta_c-2\frac{\delta_c}{\delta_u-1}\beta_u}{\left(\frac{\delta_c}{\delta_u-1}\right)^2+1}\geq\beta_c$}

\textit{(iv) For causal and causal effect regularization, denote the coefficients assigned to $C$, $U$ and $S$ as $\lambda_c$, $\lambda_u$ and $\lambda_s$, the condition is $\frac{\beta_c+\frac{\lambda_u}{\lambda_s}\delta_u^2\beta_c-\frac{\lambda_u}{\lambda_s}\delta_u\delta_c\beta_u}{\frac{\lambda_c}{\lambda_s}\delta_c^2+{\frac{\lambda_u}{\lambda_s}\delta_u^2+1}}=\beta_c$}

\textbf{Proof}:

Given $c=u=s$, the true linear relationship between $C$, $U$ and $Y$ becomes $$Y=\beta_cC+\beta_uU$$

Then the predicted values based on $C$, $S$ and $U$ are $$\hat{Y}=\hat{\beta}_cC+\hat{\beta}_uU+\hat{\beta}_sS$$

The loss function becomes:
\begin{align}
loss&=\frac{1}{n}\sum_{i=1}^{n}(Y_i-\hat{Y}_i)^2+\lambda\mathcal{R}(\hat{\beta}_c,\hat{\beta}_u,\hat{\beta}_s)\\
&=\frac{1}{n}\sum_{i=1}^{n}(\beta_cC_i+\beta_uU_i-\hat{\beta}_cC_i-\hat{\beta}_uU_i-\hat{\beta}_sS_i)^2+\lambda\mathcal{R}(\hat{\beta}_c,\hat{\beta}_u,\hat{\beta}_s)\\
&=\frac{1}{n}\sum_{i=1}^{n}((\beta_c-\hat{\beta}_c)C_i+(\beta_u-\hat{\beta}_u)U_i-\hat{\beta}_s(\delta_cC_i+\delta_uU_i))^2+\lambda\mathcal{R}(\hat{\beta}_c,\hat{\beta}_u,\hat{\beta}_s)\\
&=\frac{1}{n}\sum_{i=1}^{n}((\beta_c-\hat{\beta}_c)C_i+(\beta_u-\hat{\beta}_u)U_i-\hat{\beta}_s(\delta_cC_i+\delta_uU_i))^2+\lambda\mathcal{R}(\hat{\beta}_c,\hat{\beta}_u,\hat{\beta}_s)\\
&=\frac{1}{n}\sum_{i=1}^{n}((\beta_c-\hat{\beta}_c-\delta_c\hat{\beta}_s)C_i+(\beta_u-\hat{\beta}_u-\delta_u\hat{\beta}_s)U_i))^2+\lambda\mathcal{R}(\hat{\beta}_c,\hat{\beta}_u,\hat{\beta}_s)
\end{align}

Select $\lambda$ so that the empirical loss $\frac{1}{n}\sum_{i=1}^{n}(Y_i-\hat{Y}_i)^2$ becomes no larger after applying regularization. Because of the linear relationship between $C$, $U$ and $S$, the empirical loss can be $0$ when 
\begin{align}
    \beta_c-\hat{\beta}_c-\delta_c\hat{\beta}_s = 0\\
    \beta_u-\hat{\beta}_u-\delta_u\hat{\beta}_s = 0
\end{align}
Then
\begin{align}
    &\hat{\beta}_s = \frac{\beta_c-\hat{\beta}_c}{\delta_c}\label{beta_s_equ}\\
    &\hat{\beta}_u = \beta_u-\delta_u\frac{\beta_c-\hat{\beta}_c}{\delta_c}\label{beta_u_equ}
\end{align}

Because the empirical loss can be $0$, then for each regularization method, we can focus only on the regularization term $\mathcal{R}(\hat{\beta}_c,\hat{\beta}_u,\hat{\beta}_s)$ when $\hat{\beta}_s$ and $\hat{\beta}_u$ follow equations~\eqref{beta_s_equ} and \eqref{beta_u_equ}.

\textbf{L1 regularization}

For L1 regularization, the regularization term becomes
\begin{align}
    \mathcal{R}_{L1}(\hat{\beta}_c,\hat{\beta}_u,\hat{\beta}_s) &= |\hat{\beta}_c|+|\hat{\beta}_u|+|\hat{\beta}_s|\\
    &=|\hat{\beta}_c|+|\beta_u-\delta_u\frac{\beta_c-\hat{\beta}_c}{\delta_c}|+|\frac{\beta_c-\hat{\beta}_c}{\delta_c}|
\end{align}
Without loss of generality, assume $\hat{\beta}_c$, $\hat{\beta}_u$ and $\hat{\beta}_s$ are non-negative, then
\begin{align}
    \mathcal{R}_{L1}(\hat{\beta}_c,\hat{\beta}_u,\hat{\beta}_s)
    &=|\hat{\beta}_c|+|\beta_u-\delta_u\frac{\beta_c-\hat{\beta}_c}{\delta_c}|+|\frac{\beta_c-\hat{\beta}_c}{\delta_c}|\\
    &= \hat{\beta}_c + \beta_u-\delta_u\frac{\beta_c-\hat{\beta}_c}{\delta_c}+\frac{\beta_c-\hat{\beta}_c}{\delta_c}\\
    &=\frac{\delta_c+\delta_u-1}{\delta_c}\hat{\beta}_c + \beta_u+\frac{1-\delta_u}{\delta_c}\beta_c
\end{align}
Because $\hat{\beta}_c$, $\hat{\beta}_u$ and $\hat{\beta}_s$ are non-negative, then $\hat{\beta}_c\in[\max\{0,\frac{\delta_c}{\delta_u}\beta_u\},\beta_c]$. To make $\mathcal{R}$ achieves its minimum when $\hat{\beta}_s=0$, equivalently $\hat{\beta}_c=\beta_c$, $\frac{\delta_c+\delta_u-1}{\delta_c}\leq0$. $\square$

\textbf{EYE regularization}

For EYE regularization, the regularization term becomes
\begin{align}
    \mathcal{R}_{EYE}(\hat{\beta}_c,\hat{\beta}_u,\hat{\beta}_s) &= |\hat{\beta}_u|+|\hat{\beta}_s| + \sqrt{\left(|\hat{\beta}_u|+|\hat{\beta}_s|\right)^2+\hat{\beta}_c^2}\\
\end{align}

Similarly, assume $\hat{\beta}_c$, $\hat{\beta}_u$ and $\hat{\beta}_s$ are non-negative, then $\hat{\beta}_c\in[\max\{0,\frac{\delta_c}{\delta_u}\beta_u\},\beta_c]$ and the regularization term:
\begin{align}
     \mathcal{R}_{EYE}(\hat{\beta}_c,\hat{\beta}_u,\hat{\beta}_s) &= \hat{\beta}_u+\hat{\beta}_s + \sqrt{\left(\hat{\beta}_u+\hat{\beta}_s\right)^2+\hat{\beta}_c^2}\\
     &=\beta_u-\delta_u\frac{\beta_c-\hat{\beta}_c}{\delta_c}+\frac{\beta_c-\hat{\beta}_c}{\delta_c}+\sqrt{(\beta_u-\delta_u\frac{\beta_c-\hat{\beta}_c}{\delta_c}+\frac{\beta_c-\hat{\beta}_c}{\delta_c})^2+\hat{\beta}_c^2}\\
     \frac{\partial\mathcal{R}_{EYE}(\hat{\beta}_c,\hat{\beta}_u,\hat{\beta}_s)}{\partial\hat{\beta}_c}&=\frac{\delta_u-1}{\delta_c}+\dfrac{(\beta_u-\delta_u\frac{\beta_c-\hat{\beta}_c}{\delta_c}+\frac{\beta_c-\hat{\beta}_c}{\delta_c})(\frac{\delta_u-1}{\delta_c})+\hat{\beta}_c}{\sqrt{(\beta_u-\delta_u\frac{\beta_c-\hat{\beta}_c}{\delta_c}+\frac{\beta_c-\hat{\beta}_c}{\delta_c})^2+\hat{\beta}_c^2}}
\end{align}
Then, by setting the derivative with respect to $\hat{\beta}_c$ to $0$, we can get when the regularization term achieves minimum, which is
\begin{align}
    \hat{\beta}_c^{opt}=\frac{2\beta_c-2\frac{\delta_c}{\delta_u-1}\beta_u}{\left(\frac{\delta_c}{\delta_u-1
    }\right)^2+1}
\end{align}
Then $\mathcal{R}_{EYE}(\hat{\beta}_c,\hat{\beta}_u,\hat{\beta}_s)$ is decreasing within $[0,\hat{\beta}_c]$ and increasing within $[\hat{\beta}_c,\beta_c)$. Therefore, to make sure $\hat{\beta}_c$ achieves its minimum when $\hat{\beta}_s=0$, i.e. $\hat{\beta}_c=\beta_c$, it is required that
\begin{align}
    \hat{\beta}_c^{opt}=\frac{2\beta_c-2\frac{\delta_c}{\delta_u-1}\beta_u}{\left(\frac{\delta_c}{\delta_u-1
    }\right)^2+1}\geq\beta_c
\end{align}

\textbf{L2, causal and causal effect regularization}

For L2, causal and causal effect regularization, the regularization term becomes
\begin{align}
    \mathcal{R}_{L1}(\hat{\beta}_c,\hat{\beta}_u,\hat{\beta}_s) &=\lambda_c\hat{\beta}_c^2+\lambda_u\hat{\beta}_u^2+\lambda_s\hat{\beta}_s^2\\
    &=\lambda_c\hat{\beta}_c^2+\lambda_u\left(\beta_u-\delta_u\frac{\beta_c-\hat{\beta}_c}{\delta_c}\right)^2+\lambda_s\left(\frac{\beta_c-\hat{\beta}_c}{\delta_c}\right)^2\\
\frac{\partial \mathcal{R}_{L1}(\hat{\beta}_c,\hat{\beta}_u,\hat{\beta}_s)}{\partial \hat{\beta}_c}&=2\lambda_c\hat{\beta}_c+2\lambda_u\frac{\delta_u}{\delta_c}\left(\beta_u-\delta_u\frac{\beta_c-\hat{\beta}_c}{\delta_c}\right)-2\frac{\lambda_s}{\delta_c}\left(\frac{\beta_c-\hat{\beta}_c}{\delta_c}\right)
    % &=\left(\lambda_c+\lambda_u(\frac{\delta_u}{\delta_c})^2+\frac{\lambda_s}{\delta_c^2}\right)\hat{\beta}_c^2+2\left(\lambda_u\frac{\delta_u}{\delta_c}\beta_u-\lambda_u\beta_c(\frac{\delta_u}{\delta_c})^2-\frac{\lambda_s}{\delta_c^2}\beta_c\right)\hat{\beta}_c+\lambda_u\beta_u^2+\lambda_u(\frac{\delta_u}{\delta_c})^2\beta_c^2-2\lambda_u\beta_c\beta_u\frac{\delta_u}{\delta_c}+\frac{\lambda_s}{\delta_c^2}\beta_c^2
\end{align}
Hence, set the first derivative with respect to $\hat{\beta}_c$ to be $0$ according to KKT condition.
\begin{align}
    \hat{\beta_c}&=\dfrac{-\lambda_u\beta_u\frac{\delta_u}{\delta_c}+\lambda_u\beta_c(\frac{\delta_u}{\delta_c})^2+\frac{\lambda_s}{\delta_c^2}\beta_c}{\lambda_c+\lambda_u(\frac{\delta_u}{\delta_c})^2+\frac{\lambda_s}{\delta_c^2}}\\
    &=\dfrac{\lambda_s\beta_c+\lambda_u\beta_c\delta_u^2-\lambda_u\beta_u\delta_u\delta_c}{\lambda_c\delta_c^2+\lambda_u\delta_u^2+\lambda_s}\\
    &=\dfrac{\beta_c+\frac{\lambda_u}{\lambda_s}\delta_u^2\beta_c-\frac{\lambda_u}{\lambda_s}\delta_u\delta_c\beta_u}{\frac{\lambda_c}{\lambda_s}\delta_c^2+{\frac{\lambda_u}{\lambda_s}\delta_u^2+1}}
\end{align}
Therefore, for L2, causal and causal effect regularization, the regularization can mitigate the shortcuts only when \begin{equation}
    \hat{\beta_c}=\dfrac{\beta_c+\frac{\lambda_u}{\lambda_s}\delta_u^2\beta_c-\frac{\lambda_u}{\lambda_s}\delta_u\delta_c\beta_u}{\frac{\lambda_c}{\lambda_s}\delta_c^2+{\frac{\lambda_u}{\lambda_s}\delta_u^2+1}}=\beta_c
\end{equation}

\section{Experiments}
Comprehensive experiments are conducted on three datasets, synthetic dataset, Colored-MNIST and MultiNLI to evaluate the effectiveness of different regularization methods. The codes will be made available during the review process.

\subsection{Synthetic dataset}\label{synthetic_dataset}

For synthetic dataset, we generated two known concepts $C_1$, $C_2$ and two unknown concepts $U_1$, $U_2$ and one shortcut variable $S$. The output label $Y$ is generated by equation:
\begin{equation}
    Y=4C_1-\frac{1}{2}C_2+U_1+2U_2
\end{equation}
We design three types of shortcuts with different relationships to known and unknown concepts.
\begin{itemize}
    \item $S$ is only correlated with $C_1$ and $C_2$
    $$S=1.5C_1-0.5C_2$$
    \item $S$ is correlated with $U_1$ and $U_2$
    $$S=-0.5C_2+U_1+2U_2$$
    \item $S$ is correlated with $Y$
    $$S=Y+\epsilon,\hspace{1em}\text{$\epsilon\sim\mathcal{N}(0,1)$ is some random noise}$$
\end{itemize}

\begin{figure}[ht]
    \centering
    \includegraphics[width=\linewidth]{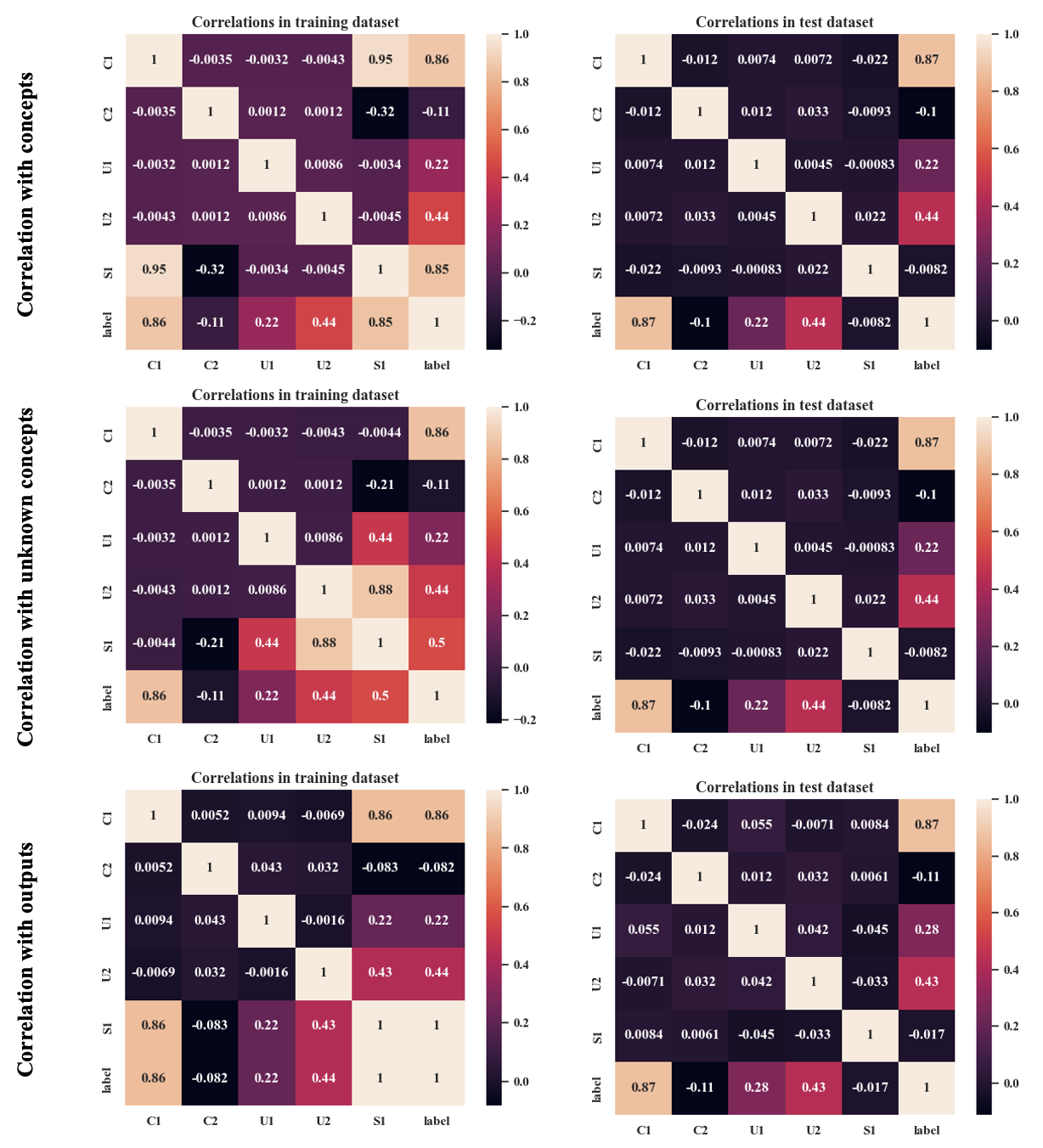}
    \caption{Correlations in the synthetic dataset for the training and test datasets. In the training dataset, the shortcut variable $S_1$ is designed to correlate with the output label by interacting with different variables, allowing the evaluation of regularization methods under varying conditions. In contrast, in the test dataset, the shortcut variable is independent of all other variables and the output label.}
    \label{fig:corr_synthetic}
\end{figure}

In the training dataset, the shortcut variable is correlated with both known and unknown concepts. However, in the test dataset, the shortcut variable is generated by randomly sampling from a standard normal distribution. The correlations within training datasets and test datasets are shown in Figure \ref{fig:corr_synthetic}. The figure illustrates that in the training datasets, the shortcut variable is designed to correlate with an increasing number of variables, while in the test dataset, it remains independent of all other variables.

\textbf{Causal effect regularization}. Regarding causal regularization, the weights corresponding to each variable are typically determined by treatment effect estimation. The performance of the treatment effect estimation can significantly impact the effectiveness of causal regularization. However, in our simulation experiments, we want to get rid of the effects of treatment estimation methods performance and only focus on the regularization itself. To achieve this, we manually assigned weights as the shortcut variables were known in our experimental setup. Specifically, we set the weights for known and unknown concepts to $1$, while assigning $0.001$ to the shortcut variables. This approach allowed us to focus solely on the behavior of the regularization.

\subsubsection{Nonlinear case}\label{nonlinear_case}
If we relax the assumption of a linear relationship (Assumption 2), we can extend regularization methods to neural network models. For multi-layer neural networks, regularization can be applied to the parameters of the first layer. For example, consider our synthetic dataset where the input data contains five variables, assume the first layer of the neural network contains $10$ neurons. The parameter matrix from the input data to the first layer would have a shape of $(10, 5)$. We can then apply regularization to these parameters, distinguishing between known and unknown concepts: let first two columns (assigned to known concepts) to be $\beta_c$ and the second two columns (assigned to unknown concepts) to be $\beta_u$.

\begin{figure}[ht]
    \centering
    \includegraphics[width=\linewidth]{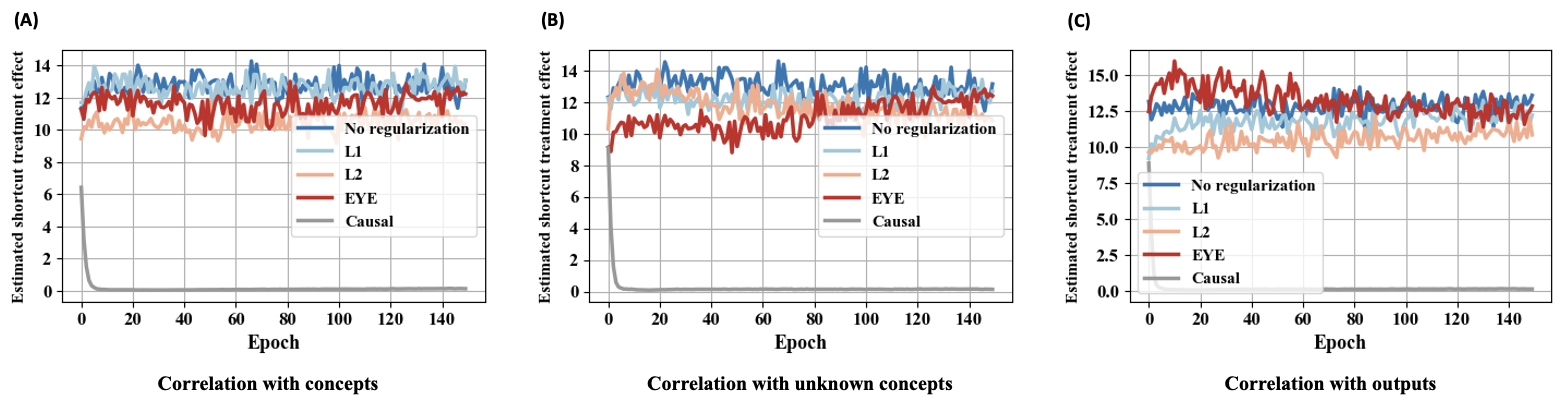}
    \caption{Treatment effect estimation of the shortcut variable along with training epoch when applying regularization terms to different shortcut types. \textbf{Neural networks might undermine the effectiveness of regularization methods.} (A) Shortcut $S$ is only correlated with $C_1$ and $C_2$. (B) $S$ is correlated with $U_1$ and $U_2$. (C) $S$ is correlated with $Y$.}
    \label{fig:te_neural_network}
\end{figure}

Since neural networks are black-box models, we cannot demonstrate whether the model mitigates shortcuts by examining the weights, as we do with linear models. Instead, we can treat neural networks as T-learners and compute the treatment effect of the shortcut variable. Ideally, if the shortcuts are mitigated, the estimated treatment effect should approach zero. The results for the synthetic datasets are presented in Figure \ref{fig:te_neural_network}.

\subsubsection{Correlation with output label}
To further evaluate the performance of various regularization methods, we calculate the correlations between predicted values and other variables, as shown in Figure \ref{fig:y_pred_correlation}.

\textbf{Correlation with the shortcut variable}: Consistent with the results in Figure \ref{fig:fail_case}, when regularization methods fail to mitigate shortcuts, the correlation between predicted values and the shortcut variable approaches 1, indicating heavy reliance on the shortcut. Moreover, as the correlation between the shortcut and other variables increases, the corresponding correlations with predicted values also rise.

\textbf{Correlation with true output values}: Similarly, the correlations between the predicted values and the true output values follow the same pattern. When the shortcuts are mitigated (i.e., correlation with the true concepts), the correlations are close to 1, indicating near-perfect predictions. However, when regularization methods fail, these correlations drop significantly, showing that reliance on shortcuts leads to poorer performance on the test dataset.

\subsubsection{Correlation with unknown concepts}\label{correlation_with_unknown_concepts}

In this experiment, the shortcut variable is generated based on $\bm{C}$ and $\bm{U}$, which makes the shortcut varibles correlated with $U$, i.e, $$S=\delta_c C+\delta_uU+\epsilon$$ where $\epsilon$ is a standard noise. Changing $\delta_u$ changes the correlation between the shortcut variable and the unknown concept. For each correlation case, we apply all regulariztion methods. The results are shown in Table \ref{table:correlation_us}.

\begin{table}[!ht]
    \centering
    \begin{tabular}{l|ccccc}
    \toprule
        Correlation between U and S & No regularization & L1 & L2 & EYE & Causal \\
    \midrule
        -0.0377 (training dataset) & 0.0 & 0.0 & 1e-05 & 1e-05 & 3e-05 \\ 
        0 (test dataset) & 0.7559 & 0.0 & 0.0001 & 0.1972 & 0.0001 \\ 
        0.0589 (training dataset) & 0.0 & 0.0 & 0.0 & 1e-05 & 3e-05 \\ 
        0 (test dataset) & 0.6589 & 0.0 & 0.1835 & 0.7066 & 0.0002 \\ 
        0.1539 (training dataset) & 1e-05 & 1e-05 & 0.0 & 0.0 & 3e-05 \\ 
        0 (test dataset) & 0.5503 & 0.0 & 0.5978 & 1.5746 & 0.0002 \\ 
        0.2448 (training dataset) & 0.0 & 0.0 & 0.0 & 0.0 & 3e-05 \\ 
        0 (test dataset) & 0.4373 & 0.0 & 1.019 & 2.7988 & 0.0002 \\
        0.3297 (training dataset) & 0.0 & 0.0 & 0.0 & 0.0 & 3e-05 \\
        0 (test dataset) & 0.3319 & 0.0 & 1.3178 & 4.3008 & 0.0002 \\
        0.4073 (training dataset) & 0.0 & 0.0 & 0.0 & 0.0 & 3e-05 \\
        0 (test dataset) & 0.2396 & 0.2395 & 1.48 & 5.8909 & 0.0002 \\
    \bottomrule
    \end{tabular}
    \caption{Mean square error of different correlations between unknown concepts (U) and shortcut variables (S).}
    \label{table:correlation_us}
\end{table}

From the Table \ref{table:correlation_us}, it is evident that as the correlation between $U$ and $S$ increases in the training dataset, the effectiveness of regularization methods in mitigating shortcuts diminishes significantly. This is reflected in the increasing discrepancy between the MSEs in the training and test datasets. These findings further demonstrate that regularization methods are likely to fail when shortcuts are highly correlated with unknown concepts. In addition, during this experiments, all the shortcuts and unknown concepts are not perfectly correlated (correlation less than 1).

% \begin{figure}
%     \centering
%     \includegraphics[width=\linewidth]{supplementary_img/y_pred_correlation.png}
%     \caption{Correlations between predicted values and all variables, including true outputs in test dataset, for different regularization methods. \textbf{
%     As shortcut mitigation becomes more effective, the correlations between predicted values and shortcuts decrease.}}
%     \label{fig:y_pred_correlation}
% \end{figure}

\subsubsection{Distribution shift}

We evaluate the effectiveness of regularization methods in managing distribution of learned weights. The primary objective, for both synthetic and real-world datasets, is to determine whether regularization can effectively constrain weights within a specific range. For instance, we assess whether regularization methods can drive the weights associated with shortcut variables toward zero.

\begin{figure}[ht!]
    \centering
    \includegraphics[width=\linewidth]{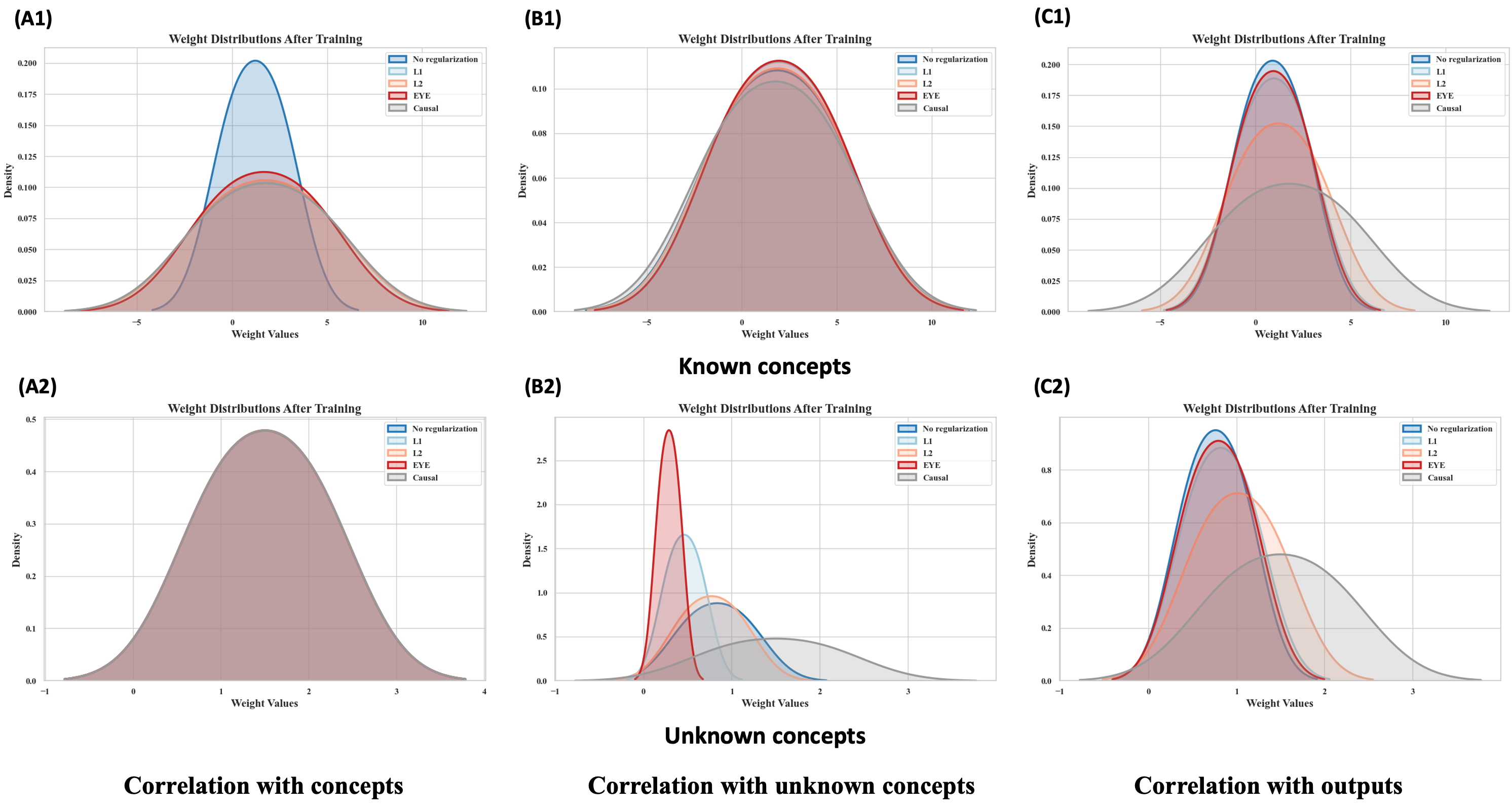}
    \caption{Distribution of learned weights across different regularization techniques for the synthetic dataset. The top row shows the distribution of weights for known concepts, while the bottom row shows the distribution for unknown concepts. Columns represent different shortcut correlations: with known concepts, unknown concepts, and the output label. \textbf{Over-regularization often happens on unknown concepts rather than known concepts}.}
    \label{fig:distribution_synthetic}
\end{figure}

\textbf{Over-regularization on known concepts is uncommon.} Figure \ref{fig:distribution_synthetic} illustrates the weight distributions for known and unknown concepts across three types of shortcut correlations. The first row shows the distributions for known concepts, while the second row focuses on unknown concepts. For known concepts (first row), when shortcuts are successfully mitigated (i.e., when the shortcut correlates only with the concepts), regularization methods broaden the range of weights assigned to the known concepts compared to no regularization, increasing their influence. However, when regularization fails to mitigate shortcuts, two scenarios arise:
\begin{enumerate}
    \item If the shortcut correlates with unknown concepts, no adjustment is needed for the weights assigned to the known concepts because the shortcuts mainly come from the unknown concepts.
    \item If the shortcut correlates with the output, only causal regularization with accurate causal effect estimations effectively broadens the weight range for the known concepts, preventing it from being concentrated near 0.
\end{enumerate}

% all regularization methods—L1, L2, EYE, and Causal—perform similarly when the shortcut is correlated with known concepts, retaining much of the information from the known variables. No regularization, however, allows for greater variability in the weights, as indicated by the broader distribution. When the shortcut is correlated with unknown concepts or the output, regularization methods still maintain the known concepts' weights.

\textbf{Over-regularization on unknown concepts is common.} The second row in Figure \ref{fig:distribution_synthetic} displays the weight distributions for unknown concepts. When the shortcut correlates with known concepts, no adjustment is needed for the weights of the unknown concepts (shown in panel A2), as they are not the source of the shortcut. However, when regularization fails to mitigate shortcuts, the shortcuts become correlated with unknown concepts in both cases. For example, in panel (B2), L1 and EYE regularization incorrectly constrain the weights of the unknown concepts to near zero, leading to over-regularization of the causal concepts. In contrast, in panel (C2), these methods do not present too much over-regularization of the unknown concepts because the correlation between the shortcut and the unknown concepts is weaker in this scenario.

% In the second row (unknown concepts), all regularization methods suppress the weights of unknown concepts, particularly when the shortcut is correlated with unknown concepts or the output. L1, L2, and EYE regularization reduce the influence of these concepts significantly. While this could be problematic, as these unknown concepts may still carry valuable information, regularization methods treat them as confounders. Causal regularization, however, preserves some of the unknown concept weights, indicating a more balanced approach in reducing shortcut reliance while maintaining potential useful information from unknown concepts.

% Overall, causal regularization is most effective at minimizing the shortcut's impact across different correlation scenarios, while L1, L2, and EYE struggle more, particularly when the shortcut is correlated with the output. The suppression of unknown concepts by these methods suggests they might over-penalize important features that are not explicitly recognized as causal, resulting in a limitation of regularization in managing complex feature interactions.

\subsection{Colored-MNIST \& MultiNLI}
\subsubsection{Data generation and shortcut allocation}\label{real_dataset}

\textbf{Colored-MNIST}

Following \cite{arjovsky2019invariant}, we first adapted the \texttt{MNIST} dataset \citep{lecun1998gradient} by 
\begin{enumerate}
    \item Binarize the labels (digits 0-4 as class $0$, 5-9 as class $1$)
    \item Set color labels $s$ equal to $y$ in the training set while reversing this in the test set. In training dataset, all the digits with class $0$ are assigned with green colors and all the digits with class $1$ are assigned with red colors. In test dataset, the assignment rule is reversed.
\end{enumerate}

Thus, image color (green or red) acts as a shortcut for predicting the binary label. Additionally, we generated an unbiased dataset with random color assignments, ensuring that both colors are evenly distributed across both label classes. As a result, the model trained on this unbiased data is free from relying on shortcuts, which is further validated by its performance on the test dataset.

Two models were then trained: one for binary label prediction and another for color prediction. The last layer of the binary label model was used to extract concepts, while the last layer of the color prediction model was used to extract shortcuts. For unknown concept extraction, we utilized a pre-trained \texttt{resnet50} from \texttt{torchvision}. All concepts, unknown concepts, and shortcuts were represented with a dimension of $50$. Since the last layer of the pre-trained \texttt{resnet50} has a dimension of $1000$, we added a fully connected layer to reduce it from $1000$ to $50$, generating unknown concepts with the desired dimension. The parameters of this fully connected layer were initialized using a uniform distribution within the range $(-1, 1)$. Because this additional layer lacks an activation function, it preserves the linearity of the unknown features during further regularization training, which does not affect the regularization performance. Regarding the weights assigned to the variables in causal effect regularization, we set the weights for known and unknown concepts to $1$, while assigning $0.001$ to the shortcut variables to remain consistent with the settings in the experiments of the synthetic datasets. This approach allowed us to focus solely on the behavior of the regularization.

\textbf{MultiNLI}

For the MultiNLI dataset \citep{williams2017broad}, the task is to classify whether a hypothesis is entailed or contradicts a premise. We follow a similar approach to the shortcut allocation used in Colored-MNIST, with the following steps:
\begin{enumerate}
    \item Exclude all records labeled as "neutral."
    \item For the training dataset, select samples where the presence of negation in the second sentence corresponds to "contradiction," and for the test dataset, select samples where negation corresponds to "entailment." This reverse correlation allows us to assess the effectiveness of shortcut mitigation more clearly.
\end{enumerate}

Therefore, having negation words in the second sentence acts as a shortcut for the classification. Similarly, we generated an unbiased dataset with random negation words assignments, ensuring that having or not having negation words are evenly distributed across both entailment and contradiction samples.

Two BERT-based models were trained: one for the entailment versus contradiction classification task, and another for detecting whether the second sentence contains negation words. The last layer of the first model was used to extract concepts, while the last layer of the second model was used to extract shortcuts, both with a dimension of 50. To extract unknown concepts, we employed a pre-trained \texttt{bert-base-uncased} model.

\textbf{MIMIC-ICU}

For the MIMIC-ICU dataset \citep{johnson2023mimic} (\url{https://physionet.org/content/mimiciv/3.1/}), the task is to predict the length of stay (LOS) in intensive care unit (ICU). For the experiments, we identified 19 variables relevant to the prediction of LOS. These variables were categorized into known concepts and unknown concepts based on their correlations with LOS. Specifically, variables with a correlation greater than 0.1 with the output LOS were classified as known concepts, while the remaining variables were considered unknown concepts. This approach was necessary because there are no universally pre-trained models tailored for extracting information from this dataset. As a result, we identified 12 known concepts and 7 unknown concepts.

For shortcut variables, we generated 10 shortcuts derived from LOS. All shortcut variables were highly correlated with LOS in the training dataset but were randomly distributed in the test dataset to simulate spurious correlations. Consistent with the methodology of our other experiments, we applied various regularization methods to train a regression model on this dataset. Model performance was evaluated using the Mean Square Error (MSE).

\begin{figure*}[ht]
    \centering
    \includegraphics[width=\linewidth]{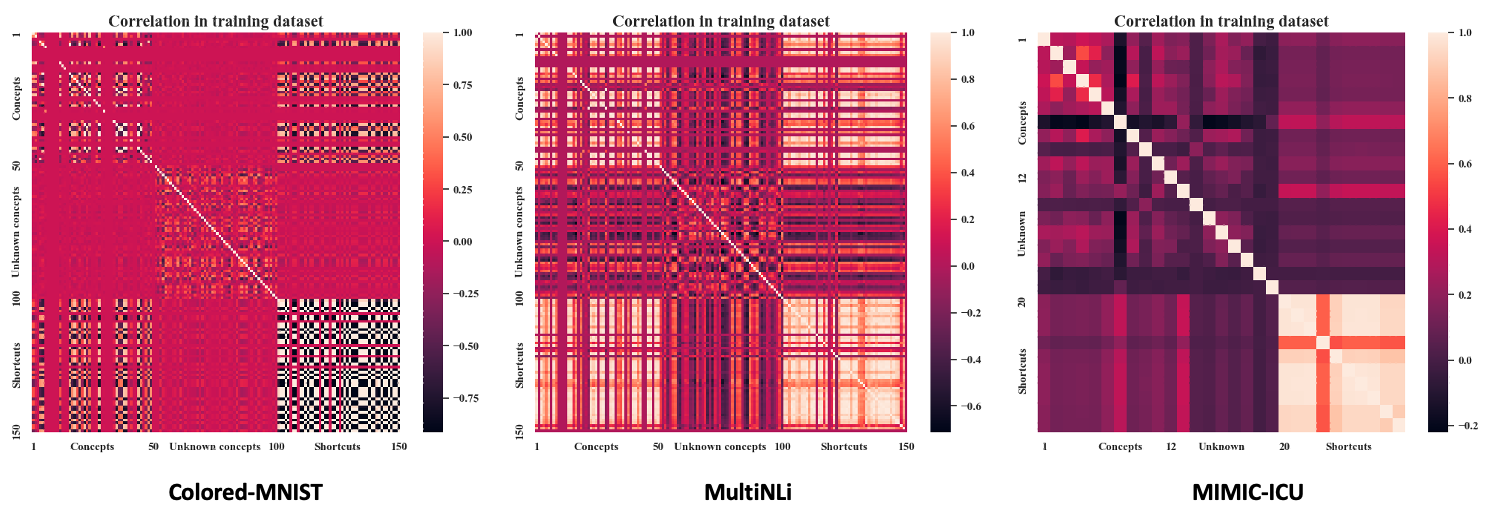}
    \caption{Correlation among the data for Colored-MNIST, MultiNLI and MIMIC-ICU datasets.}
    \label{fig:correlation_verification}
    \vspace{-0.5em}
\end{figure*}

Correlations within variables in the datasets are shown in Figure \ref{fig:correlation_verification}.

\subsubsection{Distribution shift}

\begin{figure}[ht!]
    \centering
    \includegraphics[width=\linewidth]{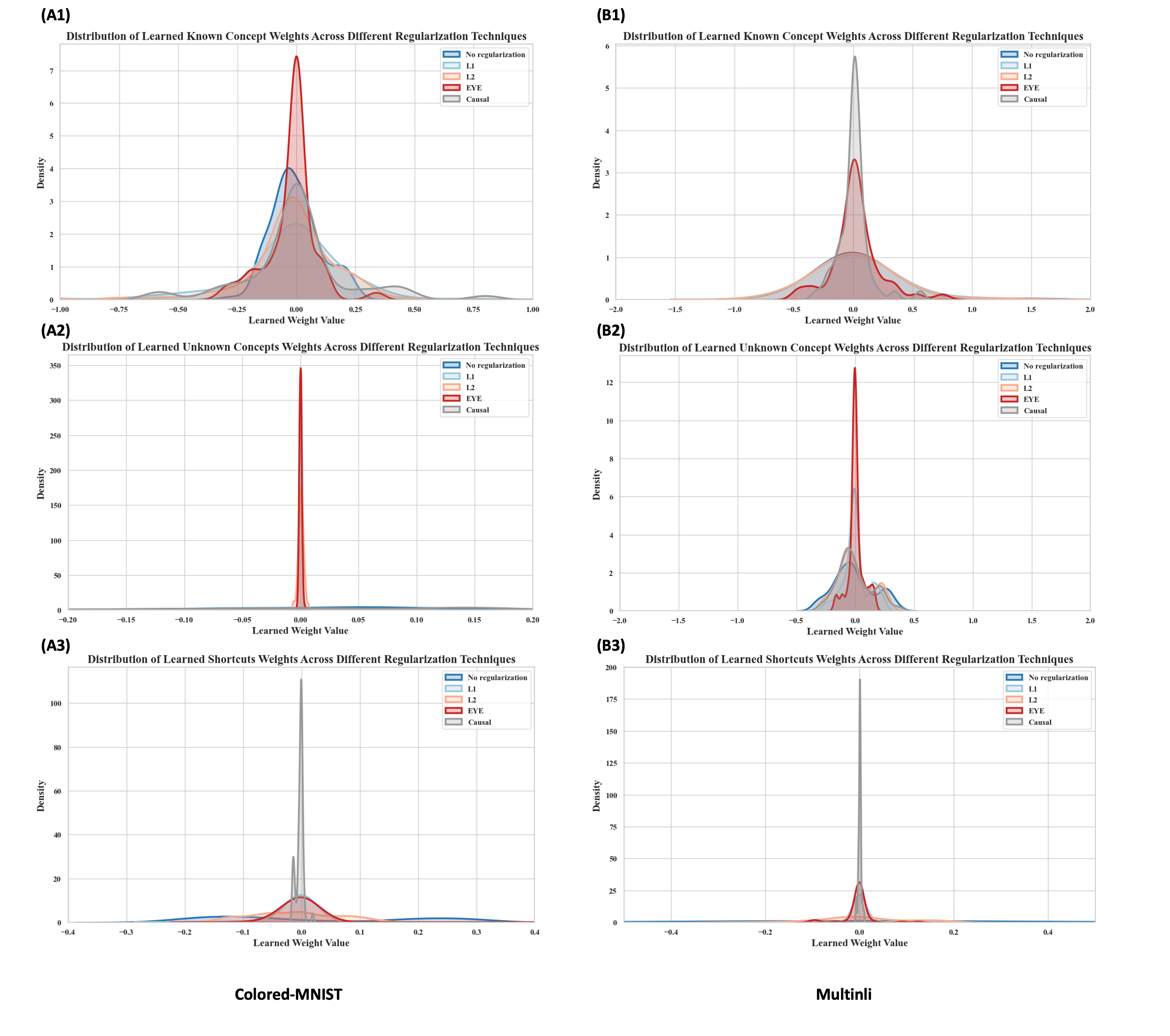}
    \caption{Distribution of learned weights across different regularization techniques for the Colored-MNIST and MultiNLI datasets. The top row shows the distribution of learned weights for known concepts, while the middle row shows the distribution for unknown concepts. The bottom row illustrates the distribution of learned weights for shortcuts. \textbf{Regularization methods tend to over regularize on unknown concepts and broaden the weight choices for the shortcut variables.}}
    \label{fig:distribution_real}
\end{figure}

For the real-world datasets, the distribution of learned weights for known concepts, unknown concepts, and shortcuts further shows the differences in how regularization methods mitigate shortcut learning. As illustrated in Figure \ref{fig:distribution_real}, the weight distributions for Colored-MNIST and MultiNLI show the effectiveness of different regularization techniques in shortcut mitigation.

For both datasets, different regularization methods show varying performances in terms of learned weight distribution. For known concepts (panel A1), EYE regularization penalizes the known concepts the most, narrowing the weight distribution to near zero. In contrast, the other three regularization methods broaden the weight distribution, encouraging the model to rely more on the known concepts. However, for unknown concepts (panel A2 and B2), all regularization methods, except causal regularization, suppress their influence on the final predictions by concentrating the weights near zero. Correspondingly, these methods allow the weights of the shortcut variables to vary, indicating a reliance on shortcuts (panel A3 and B3). 

Despite this, the test dataset's AUC performance for Colored-MNIST appears promising (Figure \ref{fig:Results}). This may be due to two factors:
\begin{itemize}
\vspace{-1em}
\setlength\itemsep{-0.2em}
    \item The known concepts provide sufficient information for accurately predicting the binary label.
    \item The task is simple enough that fluctuations in predicted probabilities caused by the use of shortcuts do not significantly affect the binary classification outcome.
\end{itemize}

Compared the AUC performance for MultiNLI (Figure \ref{fig:Results}), the performance is not as good as those for Colored-MNIST dataset, this might be because:
\begin{itemize}
\vspace{-1em}
\setlength\itemsep{-0.2em}
    \item The known and unknown concepts are highly correlated with the shortcut variables (Figure \ref{fig:correlation_verification}).
    \item The MultiNLI task is more complex than the binary classification in Colored-MNIST, meaning that the known concepts alone are insufficient for accurate classification, especially when the unknown concepts are also suppressed.
\end{itemize}

% the weight distributions of known concepts reveal that all regularization methods perform similarly in reducing shortcut reliance. However, the presence of shortcut reliance remains, as seen in the weight distribution of shortcuts, which remain relatively high even after regularization. While EYE regularization shows some marginal improvements compared to L1 and L2, none of the methods fully eliminate shortcut reliance, pointing to the need for additional strategies. Furthermore, the unknown concepts tend towards zero weight, which is concerning as they could provide valuable information that should not be dismissed as confounding by default.

% For \textbf{MultiNLI}, causal regularization shows the most improvement, reducing shortcut weights to near-zero. The weight distributions of known and unknown concepts remain relatively stable across regularization methods, but causal regularization stands out by minimizing shortcut reliance more robustly than L1, L2, and EYE, which still struggle in this regard. Similar to the Colored-MNIST results, unknown concepts are suppressed, which is a drawback. These unknown concepts could hold valuable information, and over-penalizing them might hinder the model's generalization capacity.

% \subsection{MultiNLI}

% \subsubsection{Shortcut assignment}
% \subsubsection{Distribution shift}

\vfill

\end{document}



% Supplementary material: To improve readability, you must use a single-column format for the supplementary material.
\onecolumn
\aistatstitle{Supplementary Materials:\\
Do Regularization Methods for Shortcut Mitigation Work As Intended?}

\section{Proofs for section \ref{section:theory}}

To prepare a supplementary pdf file, we ask the authors to use \texttt{aistats2025.sty} as a style file and to follow the same formatting instructions as in the main paper.
The only difference is that the supplementary material must be in a \emph{single-column} format.
You can use \texttt{supplement.tex} in our starter pack as a starting point, or append the supplementary content to the main paper and split the final PDF into two separate files.

Note that reviewers are under no obligation to examine your supplementary material.

\section{MISSING PROOFS}

The supplementary materials may contain detailed proofs of the results that are missing in the main paper.

\subsection{Proof of Lemma 3}

\textit{In this section, we present the detailed proof of Lemma 3 and then [ ... ]}

\section{ADDITIONAL EXPERIMENTS}

If you have additional experimental results, you may include them in the supplementary materials.

\subsection{The Effect of Regularization Parameter}

\textit{Our algorithm depends on the regularization parameter $\lambda$. Figure 1 below illustrates the effect of this parameter on the performance of our algorithm. As we can see, [ ... ]}

\vfill